\documentclass[article]{elsarticle}
\usepackage{lineno,hyperref}

\usepackage{graphicx}
\usepackage{latexsym}
\usepackage{color}
\usepackage{caption}
\usepackage{subcaption}
\usepackage{bm}
\usepackage{bbm}
\usepackage{amsmath,amssymb, amsthm}
\usepackage{algorithm}
\usepackage{algpseudocode}
\usepackage{color,soul}
\usepackage{graphics}
\usepackage{amsfonts}
\usepackage{setspace}   
\usepackage{wasysym}
\usepackage{multirow}

\usepackage{adjustbox}
\usepackage{graphicx}
\usepackage{multirow}
\usepackage{nomencl}
\usepackage{float}
\usepackage[table]{xcolor}
\usepackage{standalone}

\makenomenclature
\biboptions{sort&compress}

\makeatletter
\newcommand{\thickhline}{%
	\noalign {\ifnum 0=`}\fi \hrule height 1pt
	\futurelet \reserved@a \@xhline
}
\newcolumntype{"}{@{\hskip\tabcolsep\vrule width 1pt\hskip\tabcolsep}}
\makeatother

\newcolumntype{M}[1]{>{\centering\arraybackslash}m{#1}}

\usepackage{geometry}
\makeatletter
\def\ps@pprintTitle{%
	\let\@oddhead\@empty
	\let\@evenhead\@empty
	\def\@oddfoot{\reset@font\hfil\thepage\hfil}
	\let\@evenfoot\@oddfoot
}
\makeatother

\begin{document}

\begin{frontmatter}
	
	\title{Efficient Training of Physics-Informed Neural Networks via~Importance Sampling}
	
	\author{Mohammad Amin Nabian$^{1,2}$}
	\author{Rini Jasmine Gladstone$^{1}$}
	\author{Hadi Meidani$^{1,*}$}
	\address{$^1$ University of Illinois at Urbana-Champaign, Urbana, Illinois, USA. \\
	$^2$ NVIDIA, Santa Clara, California, USA. \\
    $^*$ Corresponding author (meidani@illinois.edu).}

\begin{abstract}
{Physics-Informed Neural Networks (PINNs) are a class of deep neural networks that are trained, using automatic differentiation, to compute the response of systems governed by partial differential equations (PDEs). The training of PINNs is simulation-free, and does not require any training dataset to be obtained from numerical PDE solvers. Instead, it only requires the physical problem description, including the governing laws of physics, domain geometry, initial/boundary conditions, and the material properties. This training usually involves solving a non-convex optimization problem using variants of the stochastic gradient descent method, with the gradient of the loss function approximated on a batch of collocation points, selected randomly in each iteration according to a uniform distribution. Despite the success of PINNs in accurately solving a wide variety of PDEs, the method still requires improvements in terms of computational efficiency. To this end, in this paper, we study the performance of an importance sampling approach for efficient training of PINNs. Using numerical examples together with  theoretical evidences, we show that in each training iteration, sampling the collocation points according to a distribution proportional to the loss function will improve the convergence behavior of the PINNs training. Additionally, we show that providing a piecewise constant approximation to the loss function for faster importance sampling can further improve the training efficiency. This importance sampling approach is straightforward and easy to implement in the existing PINN codes, and also does not introduce any new hyperparameter to calibrate.  The numerical examples include  elasticity, diffusion and plane stress problems, through which we numerically verify the accuracy and efficiency of the importance sampling approach compared to the predominant uniform sampling approach.}
\end{abstract}


\end{frontmatter}


\section{Introduction} \label{sec:introduction}

Physics-Informed Neural Networks (PINNs)  leverage  recent advances in deep neural networks to calculate the response of systems governed by Partial Differential Equations (PDEs).  Specifically, PINNs are trained to satisfy the governing laws of physics described in form of PDEs, as well as initial/boundary conditions and measurement data \citep{lagaris1998artificial,raissi2019physics}. In this approach, the solution to a PDE is considered to be in the form of a deep neural network (with second or higher-order differentiable nonlinearities) whose parameters are estimated by minimizing the squared residuals over specified collocation points using Automatic Differentiation (AD) \citep{baydin2018automatic} and variants of the Stochastic Gradient Descent (SGD) algorithm \citep{bottou2012stochastic}. 

The idea of using physics-informed training for  neural network solutions of differential equations  was first introduced in \cite{dissanayake1994neural,lagaris1998artificial,psichogios1992hybrid}, where neural networks solutions for initial/boundary value problems were developed. The method, however, did not gain much attention due to limitations in computational resources and optimization algorithms, until recently when researchers revisited this idea in (1) solving  challenging dynamic problems described by time-dependent nonlinear PDEs \citep{raissi2017physics,raissi2019physics}, (2) solving variants of nonlinear PDEs  \citep{berg2018unified,sirignano2017dgm,guo2019deep,weinan2018deep,goswami2019transfer,jagtap2019adaptive}, (3) data-driven discovery of PDEs (e.g. \citep{raissi2019physics,raissi2018deep,qin2018data,long2017pde}), (4) uncertainty quantification (e.g. \citep{nabian2019deep,raissi2019deep,raissi2018hiddena,zhu2019physics,meng2019composite,yang2019adversarial,kissas2019machine,xu2019neural}), (5) solving stochastic PDEs (e.g. \citep{yang2018physics,raissi2018forward,beck2018solving,weinan2017deep}), and (6) physics-driven regularization of neural network surrogates (e.g. \citep{nabian2020physics}). 

Training of PINNs usually involves solving a non-convex optimization problem using an iterative method, with the gradient of loss function approximated on a batch of collocation points, selected randomly in each iteration according to a uniform distribution. Although this iterative update is shown to result in an unbiased estimation of the gradient with bounded variance \citep{bottou2010large}, such batch selection may seem to be na\"ive in terms of efficiency. In a given iteration, such batch selection may result in computing the gradient at a number of collocation points at which the approximate solution already satisfies the differential operator to a satisfactory extent relative to other points.  As a result, little or no gradient information will be obtained which can delay the convergence. Alternatively, by following  an importance sampling \citep{press2007numerical}  scheme, in each iteration we can select a batch of collocation points that can offer more gradient information for accelerated convergence.

The performance of implementing an importance sampling-based training  has recently been evaluated on classification tasks using convolutional neural networks and recurrent neural networks \citep{katharopoulos2018not,katharopoulos2017biased,alain2015variance}, where the authors  provided  theoretical and numerical evidences showing that the training convergence speed can be maximized if, at each training iteration, samples from the training images or texts are drawn according to a proposal distribution that is proportional to the 2-norm of loss gradient with respect to model parameters. Further, it has been illustrated that computing such proposal distributions can be computationally expensive, and the authors used an approximate proposal distribution proportional to the loss function itself to improve the computational efficiency. Finally, the performance of such importance sampling approach is evaluated for image classification and  language modeling tasks.

Our contribution in this paper is twofold. First, we borrow the theoretical findings in \cite{katharopoulos2018not,katharopoulos2017biased} to propose an efficient approach for accelerated training of PINNs based on importance sampling. To the authors' knowledge, this is the first time that an importance sampling scheme is used for training of PINNs. This can be an important step toward improving the computational efficiency of PINNs compared to their traditional numerical counterparts, i.e. Finite Difference, Finite Element, and Finite Volume methods. Second, we show how a piece-wise constant approximation to the loss function, using nearest neighbor search \citep{marsland2014machine} or Voronoi tessellation \citep{aurenhammer1991voronoi}, can be used to approximate the proposal distribution to further improve the convergence behavior of PINNs training. The proposed importance sampling approach is straightforward and can be easily applied to the existing PINN codes by modifying only a few lines of the code. Furthermore, no new hyperparameters are introduced in the proposed approach.

The remainder of this paper is organized as follows. A theoretical background on physics-informed neural networks is presented in Section 2. Our proposed importance sampling approach for training of PINNs is then introduced in Section 3. Section 4 includes three numerical examples, on which the performance of the proposed importance sampling approach is evaluated. Finally, Section 5 concludes the paper.

\
\section{Deep Learning of Differential Equations } \label{sec:DNN}

\subsection{Feed-Forward Fully-Connected Deep Neural Networks}\label{DNN}
A basic architecture for deep neural networks is the feed-forward fully-connected deep neural network, which will be explained here (a more detailed introduction can be found in \cite{lecun2015deep,goodfellow2016deep}). Given the $d$-dimensional row vector $\bm{x} \in \mathbb{R}^{d}$ as model input, the $k$-dimensional output of a standard single hidden layer neural network is in the form of
\begin{equation} \label{OHL-NN}
\bm{y} = \sigma (\bm{x} \bm{W}_{1}+\bm{b}_{1}   ) \bm{W}_{2}+\bm{b}_{2},
\end{equation}
in which $\bm{W}_{1}$ and $\bm{W}_{2}$ are weight matrices of size $d\times q$ and $q\times k$, and $\bm{b}_{1}$ and $\bm{b}_{2}$ are bias vectors of size $1\times q$ and $1\times k$, respectively. The function $\sigma( \cdot  )$ is an element-wise non-linear model, known as the activation function. In deep neural networks,for each additional hidden layer, a new set of weight matrix and biases is added to Equation (\ref{OHL-NN}). Popular choices of activation functions include Sigmoid, hyperbolic tangent (Tanh), Rectified Linear Unit (ReLU), and Sine functions.

The model parameters are estimated according to
\begin{equation} \label{minimize_loss}
( \bm{W}_{1}^{*},\bm{W}_{2}^{*},\cdots,\bm{b}_{1}^{*},\bm{b}_{2}^{*},\cdots  )=\underset{{( \bm{W}_{1},\cdots,\bm{b}_{1}\cdots  )}}{\operatorname{argmin}} J(\bm{\theta};\bm{X},\bm{Y}),
\end{equation}
where $\bm{\theta}=\{ \bm{W}_{1},\bm{W}_{2},\cdots,\bm{b}_{1},\bm{b}_{2},\cdots  \}$is the set of model parameters (i.e. weights and biases). This optimization is performed iteratively using Stochastic Gradient Descent (SGD) and its variants \citep{bottou2012stochastic,kingma2014adam,duchi2011adaptive,zeiler2012adadelta,sutskever2013importance}. Specifically, at the $i^{th}$ iteration, the model parameters are updated according to
\begin{equation} \label{descent step}
\bm{\theta}^{(i+1)} = \bm{\theta}^{(i)} - \eta^{(i)} \nabla_{\bm{\theta}}J(\bm{\theta}^{(i)}; \bm{X},\bm{Y}),
\end{equation}
where $\eta^{(i)}$ is the step size in the $i^{\textit{th}}$ iteration. The gradient of loss function with respect to model parameters $\nabla_{\bm{\theta}} J$ is usually computed using \emph{backpropagation} \citep{lecun2015deep}, which is a special case of the more general technique called reverse-mode automatic differentiation \citep{baydin2018automatic}. In simplest terms, in backpropagation, the required gradient information is obtained by the backward propagation of the sensitivity of objective value at the output, utilizing the chain rule successively to compute partial derivatives of the objective with respect to each weight \citep{baydin2018automatic}. In other words, the gradient of  last layer is calculated first and the gradient of  first layer is calculated last. Partial gradient computations for one layer are reused in the gradient computations for the foregoing layers. This backward flow of information facilitates efficient computation of the gradient at each layer of the deep neural network \citep{lecun2015deep}. It is important to note that automatic (or reverse automatic) differentiation is different from symbolic or numerical differentiation, which are two common alternatives for computing derivatives \citep{baydin2018automatic}. Detailed discussions about the backpropagation algorithm can be found in  \cite{goodfellow2016deep,lecun2015deep,baydin2018automatic}.

\subsection{Physics-Informed Neural Networks} \label{PINN}

Physics-informed neural networks are a class of deep neural networks that are used to calculate the approximate solution $u(t,\bm{x}; \bm{\theta} )$  for the following generic differential equation

\begin{equation}\label{eqn:pde}
\begin{aligned}
\mathcal{N}( t,\bm{x} ; u (t,\bm{x}; \bm{\theta} ) ) =0, \; \; \; \; & t \in [ 0,T ], \bm{x}\in \mathcal{D}, \\
\mathcal{I}( \bm{x}; u (0,\bm{x}; \bm{\theta} ) )=0, \; \; \; \; & \bm{x}\in \mathcal{D},   \\
\mathcal{B}( t,\bm{x} ; u (t,\bm{x}; \bm{\theta} ) )=0, \; \; \; \;  & t \in [ 0,T ], \bm{x}\in {\partial \mathcal{D}}, 
\end{aligned}
\end{equation}
where $\bm \theta$ include the parameters of the  function form of the solution,   $\mathcal{N}(\cdot)$ is a general differential operator that may consist of time derivatives, spatial derivatives, and linear and nonlinear terms, and $\bm{x}$ is a position vector defined on a bounded continuous spatial domain $\mathcal{D} \subseteq \mathbb{R}^D , D \in \left \{ 1,2,3 \right \} $ with boundary ${\partial \mathcal{D}}$. Also, $\mathcal{I}(\cdot)$ and $\mathcal{B}(\cdot)$ denote, respectively, the initial and boundary conditions and may consist of differential, linear, or nonlinear operators.

In order to calculate the solution, i.e. calculate the parameters $\bm{\theta}$, let us consider the following non-negative residuals, defined over the entire spatial and temporal domains 
\begin{equation}\label{eqn:l2-redidual}
\begin{aligned}
r_\mathcal{N} (  \bm{\theta}  ) &=\int_{\left[ 0,T \right] \times \mathcal{D} }( \mathcal{N} (  t,\bm{x}; \bm{\theta} ) )^2 d\!t \, d\! \bm{x},   \\
r_\mathcal{I} (  \bm{\theta}  ) &=\int_{\mathcal{D} }( \mathcal{I} (  \bm{x}, \bm{p}; \bm{\theta} ) )^2 d\!\bm{x},   \\
r_\mathcal{B} (  \bm{\theta}  )&=\int_{\left[ 0,T \right] \times {\partial \mathcal{D}}} ( \mathcal{B} (  t,\bm{x}; \bm{\theta} ) )^2 d\!t \, d\!\bm{x}.   
\end{aligned}
\end{equation}
The optimal  parameters $\bm{\theta^*}$ can then be calculated according to 
\begin{equation} \label{eqn:theta-star}
\begin{aligned}
\bm{\theta^*}=\underset{{ \bm{\theta} }}{\operatorname{argmin}}\, r_\mathcal{N}( \bm{\theta}  ),   \\
\text{s.t.} \quad r_\mathcal{I} (  \bm{\theta}  )=0, \, r_\mathcal{B} (  \bm{\theta}  )=0.
\end{aligned}
\end{equation}
Therefore, the solution to the differential equation defined in Equation \ref{eqn:pde} is reduced to an optimization problem, where initial and boundary conditions can   be viewed as constraints.  This constrained optimization can be reformulated as an unconstrained optimization with a modified loss function that also accommodates the constraints. The predominant approach to do so is the soft assignment of constraints, where the constraints are translated into additive penalty terms in the loss function (see e.g. \cite{sirignano2017dgm}). Other approaches also exist (e.g. hard assignment of constraints \citep{lagaris1998artificial}, and the unified approach \citep{berg2018unified}), but are not discussed here for brevity.

Let us denote the solution obtained by a PINN by $\tilde{u}(t,\bm{x}; \bm{\theta} )$. The inputs to this deep neural network are realizations from $t$ and $\bm{x}$. With soft assignment of constraints, we solve the following unconstrained optimization problem
\begin{equation} \label{loss}
\bm{\theta^*}=\underset{{ \bm{\theta} }}{\operatorname{argmin}}\, \underbrace{r_\mathcal{N}(  \bm{\theta} ) +\lambda_1 r_\mathcal{I}  (  \bm{\theta}   )+\lambda_2 r_\mathcal{B}  (  \bm{\theta}   )}_{J(  \bm{\theta} )},
\end{equation}
in which $\lambda_1$ and $\lambda_2$ are weight parameters, analogous to collocation finite element method in which weights are used to adjust the relative importance of each residual term \citep{bochev2006least}.

To solve this unconstrained optimization problem, mini-batch SGD optimization algorithms \citep{ruder2016overview} are used. In each iteration of a mini-batch SGD algorithm, the gradient of loss function is approximated through backpropagation using a batch of points of size $m$ in the input space, based on which the neural network parameters are updated. This iterative update is shown to result in an unbiased estimation of the gradient, with bounded variance \citep{bottou2010large}. Specifically, in the $i^{\textit{th}}$ iteration, we select a subset of collocation points uniformly drawn in $\left[ 0,T \right], \mathcal{D}, {\partial \mathcal{D}}$.  and compute the loss function as
\begin{equation} \label{loss-approximate}
\begin{split}
& J(\bm{\theta})  \approx  \frac{1}{m}\sum_{j \in M^{(i)}} J(\bm \theta; \bm x_j ) = \frac{1}{m}\sum_{j \in M^{(i)}} 
\bigg[ \left[\mathcal{N}\left( t_j,\bm{x}_j; \tilde{u}(t_j,\bm{x}_j;\bm{\theta} ) \right) \right]^2  \\ &+
\lambda_1 \left[\mathcal{I} \left( \bm{x}_j; \tilde{u}(0,\bm{x}_j;\bm{\theta} ) \right) \right]^2 
+\lambda_2 \left[\mathcal{B} \left( t_j,\underbar{$\bm{x}$}_j ; \tilde{u}(t_j,\underbar{$\bm{x}$}_j ;\bm{\theta} ) \right) \right]^2 \bigg],
\end{split}
\end{equation}
where $M^{(i)}$ is the set of indices of  selected collocation points at iteration $i$ with $|M^{(i)}|=m$, $J(\bm \theta; \bm x_j )$ is the per-sample loss evaluated at the $j$th collocation point, and $\{\underbar{$\bm{x}$}_j\}$ denotes the boundary collocation points.The model parameters are updated according to

\begin{equation} \label{descent step}
\bm{\theta}^{(i+1)} = \bm{\theta}^{(i)} - \eta^{(i)} \nabla_{\bm{\theta}}{J}(\bm{\theta}^{(i)}),
\end{equation}
where $\nabla_{\bm{\theta}} J$ is calculated using backpropagation \citep{baydin2018automatic}. Algorithm \ref{Algorithm1} summarizes the steps for training of a physics-informed neural network:

\begin{algorithm}[h]
	\caption{Training of the physics-informed neural networks}\label{Algorithm1}
	\begin{algorithmic}[1]
		\State Generate $N$ collocation points $\{ t_j,\bm{x}_j \}_{j=1}^{N}$ sampled from $[0,T] \times \mathcal{D}$, and $N$ boundary points $\{\underbar{$\bm{x}$}_j\}_{j=1}^{N}$ sampled from ${\partial \mathcal{D}}$.
		\State Set the model architecture (number of layers, dimensionality of each layer, and nonlinearities). Also specify optimizer hyper-parameters, $\lambda_1, \lambda_2$, batch size $m$, and error tolerance $\epsilon$.
		\State Initialize model parameters $\bm{\theta}^{(0)}$.
		\While {$J(\bm{\theta}^{(i)})>\epsilon$ }
		\State Randomly select a batch of $m$ points  out of the $N$ collocation points according to a uniform distribution.
		\State Take a descent step  $\bm{\theta}^{(i+1)} = \bm{\theta}^{(i)} - \eta^{(i)} \nabla_{\bm{\theta}}J(\bm{\theta}^{(i)})$.
		\EndWhile
		
	\end{algorithmic}
\end{algorithm}

Collocation points may be generated according to a uniform distribution, or alternatively according to a low-discrepancy sequence generator algorithm. Low-discrepancy sequences provide a means to generate quasi-random numbers with a high level of uniformity. Among the popular low-discrepancy sequences are the generalized Halton sequences \citep{halton1960efficiency,faure2009generalized}, the Sobol sequences \citep{sobol1967distribution,joe2008constructing}, and the Hammersley sets \citep{hammersley1960monte,hammersley2013monte}.

\
\section{Importance Sampling for Training of PINNs} \label{importance}
Consider the following parameter estimation problem,
\begin{align}
\begin{split}
\bm \theta^* &= \underset{{ \bm{\theta} }}{\operatorname{argmin}} \ \mathbb{E}_f \left[ J(\bm \theta) \right] \\
&=  \underset{{ \bm{\theta} }}{\operatorname{argmin}} \ \frac{1}{N} \sum_{j=1}^{N} J (\bm \theta; \bm x_j ),  \quad \bm x_j \sim {f}(\bm x),
\label{IM_integral}
\end{split}
\end{align}
where $f(\bm x)$ is the sampling distribution for the training points in the physical domain $\bm x \in \mathcal{D}$. The typical choice for this sampling distribution is the uniform distribution defined over the physical domain, i.e. $\mathcal{U}(\mathcal{D})$. In an importance sampling approach, we seek to draw training samples from an alternative sampling distribution, denoted by $q(\bm x)$, and instead estimate the neural network parameters according to 
\begin{equation}\label{IM_unbiased}
\bm \theta^*  \approx \underset{{ \bm{\theta} }}{\operatorname{argmin}} \ \frac{1}{N} \sum_{j=1}^{N} \frac{f(\bm x_j)}{q(\bm x_j)} J (\bm \theta; \bm x_j ),  \quad \bm x_j \sim q(\bm x).
\end{equation}

In this work, we effectively implement a discrete sampling scheme, by turning the continuous domain $\mathcal{D}$ into a discrete set of $N$ sample locations uniformly selected in $\mathcal{D}$, with $N>>1$. Thus, instead of the sampling density functions $f(\bm x_j)$ and $q(\bm x_j)$, we will work with discrete distributions $\{f_j \}_{j=1}^N$ and $\{q_j \}_{j=1}^N$, respectively, with $f_j=\frac{1}{N}$ at any candidate point $j$.

In order to build the corresponding SGD, for the sake of brevity, let us first consider no mini batch, i.e. $m=1$, that is
\begin{equation} \label{newdescent}
\bm{\theta}^{(i+1)} = \bm{\theta}^{(i)} - \eta^{(i)}  \underbrace{\nabla_{\bm{\theta}} \left( \frac{f_i}{q_i} J (\bm \theta^{(i)}; \bm{x}_i ) \right)}_{\bm G^{(i)}},
\end{equation}

Our objective in this work is to design a training scheme with a sampling distribution $q$ that can accelerate the convergence of Eq.~\ref{newdescent}. Authors in \cite{katharopoulos2018not} considered the following definition for convergence speed
\begin{equation}\label{convgspeed}
S^{(i)} = -{\mathbb{E}}_{f} \left[  ||\bm{\theta}^{(i+1)}-\bm{\theta}^{*} ||_2^2 - ||\bm{\theta}^{(i)}-\bm{\theta}^{*} ||_2^2  \right],
\end{equation}
and showed that
\begin{align}
\begin{split}
S^{(i)} & = 2 \eta \left( \bm{\theta}^{(i)}-\bm{\theta}^{*} \right) \mathbb{E}_{f}\left[ \bm{G}^{(i)} \right] - \eta^2 \mathbb{E}_{f}\left[ \bm{G}^{(i)} \right]^T \mathbb{E}_{f}\left[ \bm{G}^{(i)} \right] \\ 
& - \eta^2 \mathrm{Tr}\left( \mathbb{V}_{f}\left[ \bm{G}^{(i)} \right] \right). 
\end{split}
\end{align}
It was then concluded that the  convergence can be accelerated by sampling the input variables from a distribution that minimizes $\mathrm{Tr}\left( \mathbb{V}_{P}\left[ \bm{G}^{(i)} \right] \right)$ \citep{katharopoulos2018not}.

It was shown in \cite{katharopoulos2018not,alain2015variance} that this term is minimized if training samples are selected according to $q^* \propto \left \Vert  \nabla_{\bm{\theta}}J(\bm{\theta}^{(i)}) \right \Vert_2$. In the case of mini-batch SGD with batches of size $m$, this was effectively done by calculating the sampling distributions according to 
\begin{equation} \label{sampling-gradient}
q_j^{(i)} = \frac{\left \Vert \nabla_{\bm{\theta}}{J}(\bm{\theta}^{(i)};\bm x_j) \right \Vert _2}{\sum_{j=1}^{N}\left \Vert \nabla_{\bm{\theta}}{J}(\bm{\theta}^{(i)};\bm x_j) \right \Vert _2}, \;\;\; \forall j \in \{1, \cdots, N\},
\end{equation}
and selecting the mini-batch sample set $M^{(i)}$, with $|M^{(i)}|=m$, by sampling $m$ indices from a multinomial with  probabilities $\bm p^{(i)}=\{q_1^{(i)} ,\cdots, q_N^{(i)}\}$. In order to get an unbiased estimate of the gradient $\nabla_{\bm{\theta}} J(\bm{\theta})$, following Equations \ref{loss-approximate}, \ref{IM_unbiased}, the mini-batch gradient descent update rule will then be given by \citep{katharopoulos2018not,alain2015variance}


\begin{equation}\label{update_importance}
\bm{\theta}^{(i+1)} = \bm{\theta}^{(i)} - \frac{\eta^{(i)}}{m} \sum_{j \in M} \frac{1}{N q_j^{(i)}} \nabla_{\bm{\theta}}J(\bm{\theta}^{(i)};\bm x_j).
\end{equation}

This derivation  provides a theoretical evidence that training of PINNs can be accelerated using an importance sampling approach where training samples are obtained from  a distribution proportional to the 2-norm of the gradient of loss function with respect to model parameters. However, computing the 2-norm of this gradient for all of the collocation point at each iteration requires extra backpropagations through the computational graph, which can be computationally expensive. To alleviate this issue, it was shown theoretically and numerically in \cite{katharopoulos2017biased} that a linear transformation of the loss value at a training example is always greater than the 2-norm of loss gradient at that example, and that the ordering of collocation points according to their gradient norm is consistent with their ordering according to their loss value. Thus, one can  use  the loss value, instead of the gradient value, as the importance metric, according to

\begin{equation}\label{approx.q}
q_j^{(i)} \approx \frac{ J(\bm{\theta}^{(i)};\bm x_j)}{\sum_{j=1}^{N} J(\bm{\theta}^{(i)};\bm x_j) }, \;\;\; \forall j \in \{1, \cdots, N\}.
\end{equation}
Specifically, using this proposal distribution, one can select  $m$ mini-batch samples as explained earlier together with the gradient descent rule of Equation \ref{update_importance} to update the model parameters.

\begin{figure}
	\begin{center}
		\includegraphics[width=0.65\linewidth]{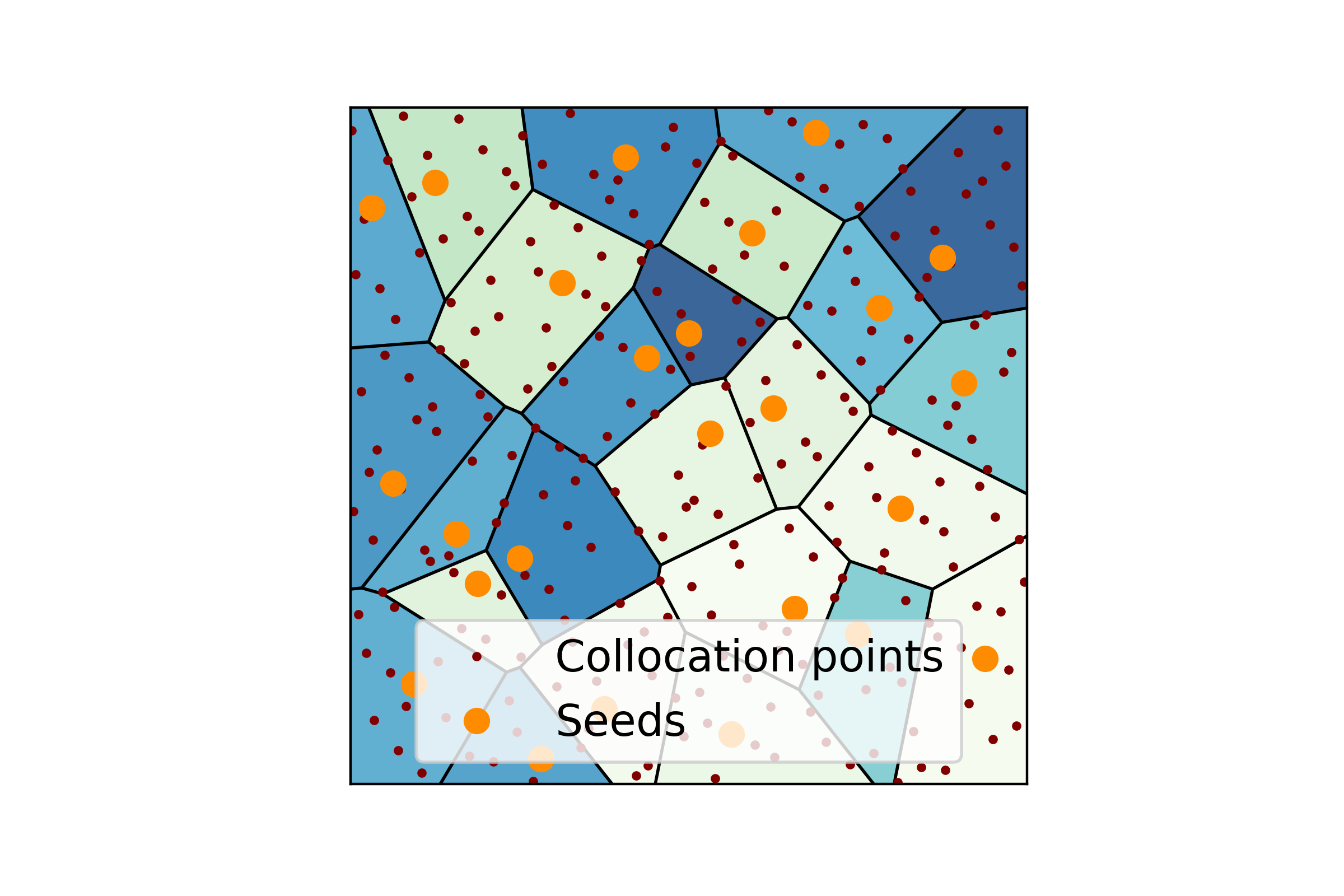}
		\caption{Piecewise constant approximation of loss on a sample 2D domain. Within each cell, the loss is evaluated only at the course-level point and is assigned to the neighboring collocation points.} 
		\label{voronoi}
	\end{center}
\end{figure}

Although evaluation of the loss function is computationally less expensive compared to that of the gradient, such evaluation for the entire set of collocation points in each iteration can  still be very expensive. To alleviate this, we propose  a piece-wise constant approximation to the loss function. That is, instead of evaluating the loss function at every collocation point, we evaluate the loss only at a subset of points, hereinafter referred to as  ``seeds", denoted by $\{\bm x_s \}_{s=1}^S$, with $S<N$. Next, using a nearest neighbor search algorithm \citep{marsland2014machine}, for each collocation point $j$, we identified the nearest seed $s=\rho(j)$, and set the loss value at that collocation point equal to the loss at the nearest seed, that is  $ J(\cdot;\bm x_j) := J(\cdot ;\bm x_{\rho(j)})$.  This is equivalent to generating a Voronoi tesselation \citep{aurenhammer1991voronoi} using the seeds, and using a constant approximation for loss within each Voronoi cell, as shown in Figure \ref{voronoi}. It will be shown in the numerical examples that such piecewise constant  approximation provides improved computational efficiency compared to the case where the loss function is evaluated for the entire collocation points.

Algorithm \ref{Algorithm2} summarizes the steps for the proposed importance sampling method for efficient training of a PINNs.

\begin{algorithm}[h]
	\caption{Efficient training of PINNs via importance sampling}\label{Algorithm2}
	\begin{algorithmic}[1]
		\State Generate $N$ collocation points $\{ t_j,\bm{x}_j \}_{j=1}^{N}$ sampled from $[0,T] \times \mathcal{D}$, $n$ boundary points $\{\underbar{$\bm{x}$}_j\}_{j=1}^{N}$ sampled from ${\partial \mathcal{D}}$, and $S$ seeds $\{ t_s,\bm{x}_s \}_{s=1}^{S}$ sampled from $[0,T] \times \mathcal{D}$.
		\State For each collocation point, find the nearest seed.
		\State Set the model architecture (number of layers, dimensionality of each layer, and nonlinearities). Also specify optimizer hyper-parameters, $\lambda_1, \lambda_2$, batch size $m$, and error tolerance $\epsilon$.
		\State Initialize model parameters $\bm{\theta}^{(0)}$.
		\While {$J( \bm{\theta} )>\epsilon$ }
		\State Compute the loss value at each seed $\{J(\bm{\theta}^{(i)}; \bm x_s)\}_{s=1}^S$.
		\State Compute $\tilde{q}_j^{(i)} = {  J(\bm{\theta}^{(i)}; \bm x_{\rho(j)}) }/{\sum_{j=1}^{N} J(\bm{\theta}^{(i)}; \bm x_{\rho(j)}) }$ for \\ 
		$\forall j \in \{1, \cdots, N\}$.
		\State Select a batch of collocation points according to a multinomial with $\bm p^{(i)}=\{\tilde{q}_1^{(i)}, \cdots, \tilde{q}_N^{(i)}\}$.
		\State Take a step  $\bm{\theta}^{(i+1)} = \bm{\theta}^{(i)} - \frac{\eta^{(i)}}{mN} \sum_{j \in M^{(i)}} \frac{1}{\tilde{q}_j^{(i)}} \nabla_{\bm{\theta}}J(\bm{\theta}^{(i)};\bm x_j)$.
		\EndWhile
		
	\end{algorithmic}
\end{algorithm}

\section{Numerical Examples} \label{examples}

In this section, we numerically demonstrate the performance of the proposed importance sampling approach, with piecewise constant (PWC) loss approximation, in solving sample PDEs. In the first example, the proposed approach is applied to solve an elasticity problem on an irregular plate. Next, a plane stress problem is solved on a 2D plate with three holes using the proposed approach. In the final example, a steady diffusion example is considered. Training is performed using TensorFlow \citep{abadi2016tensorflow} on a NVIDIA Tesla P100-PCIE-16GB GPU. The Adam optimization algorithm \citep{kingma2014adam} is used to find the optimal neural network parameters, with $\beta_1$ (the exponential decay rate for the first moment estimates) and $\beta_2$ (the exponential decay rate for the second moment estimates) set to 0.9 and 0.999, respectively.

\subsection{A two-dimensional isotropic elasticity problem} \label{example1}
In the first example, we consider the governing equations of the displacement of a two-dimensional isotropic elastic structure \citep{chikazawa2001particle} as follows

\begin{equation}\label{E1_PDE}
\resizebox{0.65\hsize}{!}{$
\begin{split}
\mathcal{N}_1 (x,y,u,v) = & (\lambda+\mu) \frac{\partial}{\partial x} \left( \frac{\partial u}{\partial x} + \frac{\partial v}{\partial y} \right) +\mu \left( \frac{\partial^2 u}{\partial x^2} + \frac{\partial^2 u}{\partial y^2} \right) + f_x, \\
\mathcal{N}_2 (x,y,u,v) = & (\lambda+\mu) \frac{\partial}{\partial y} \left( \frac{\partial u}{\partial x} + \frac{\partial v}{\partial y} \right) +\mu \left( \frac{\partial^2 v}{\partial x^2} + \frac{\partial^2 v}{\partial y^2} \right) + f_y,
\end{split}
$}
\end{equation}
where $u$ and $v$ denote the displacement along the x- and y-axes, and $f_x$ and $f_y$  the external force terms along the x- and y-axes, respectively. The constants $\lambda$ and $\mu$ are defined as

\begin{equation}
\lambda = \frac{\nu E}{(1+\nu)(1-\nu)},
\end{equation}
\begin{equation}
\mu= \frac{E}{2(1+\nu)},
\end{equation}
where $\nu$ and $E$ are the Poisson's ratio and the Young's modulus of the structure.

\begin{figure}
	\begin{center}
		\begin{subfigure}{0.48\linewidth}
			\includegraphics[width=0.99\linewidth]{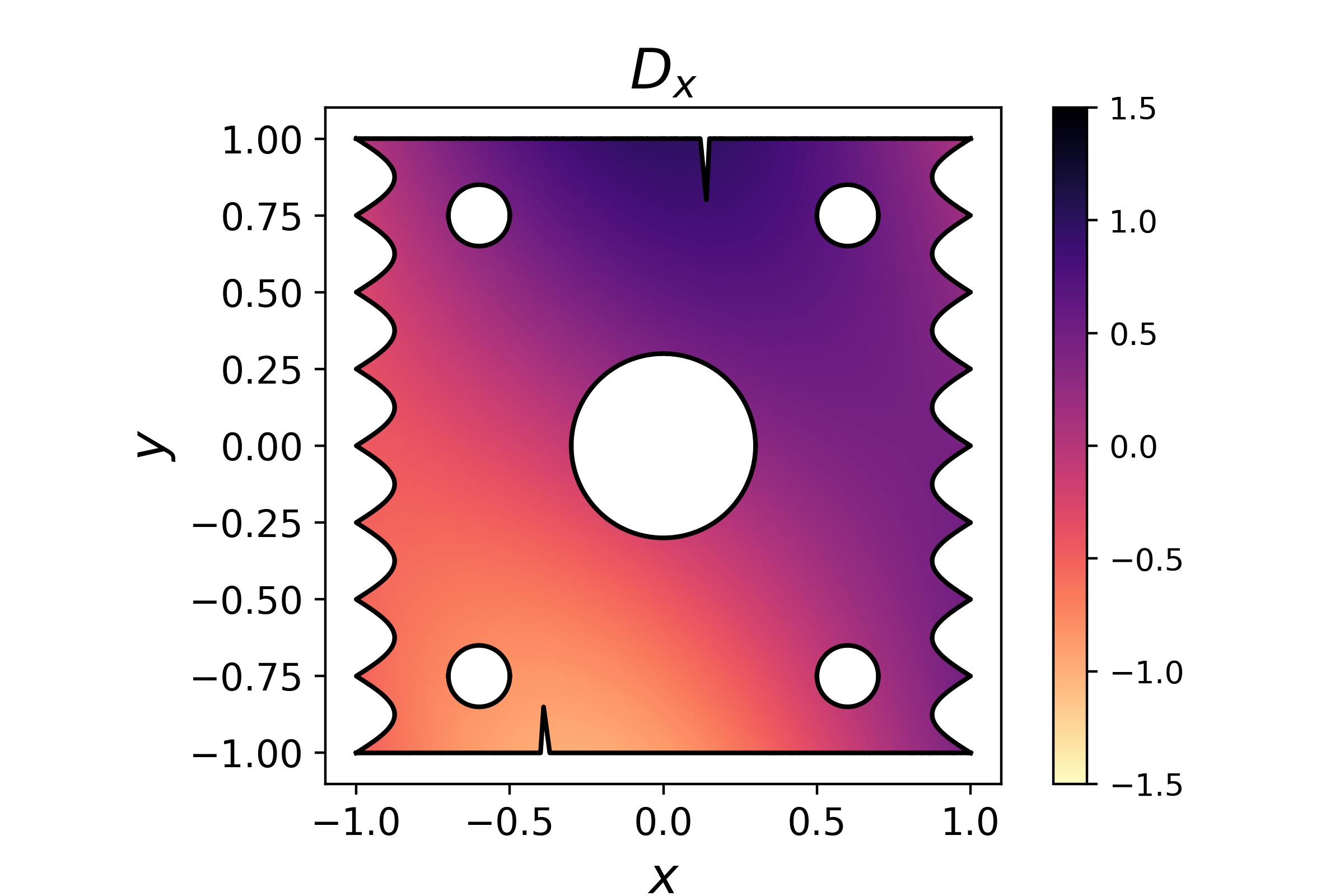}
			\caption{} 
		\end{subfigure}
		\quad
		\begin{subfigure}{0.48 \linewidth}
			\includegraphics[width=0.99\linewidth]{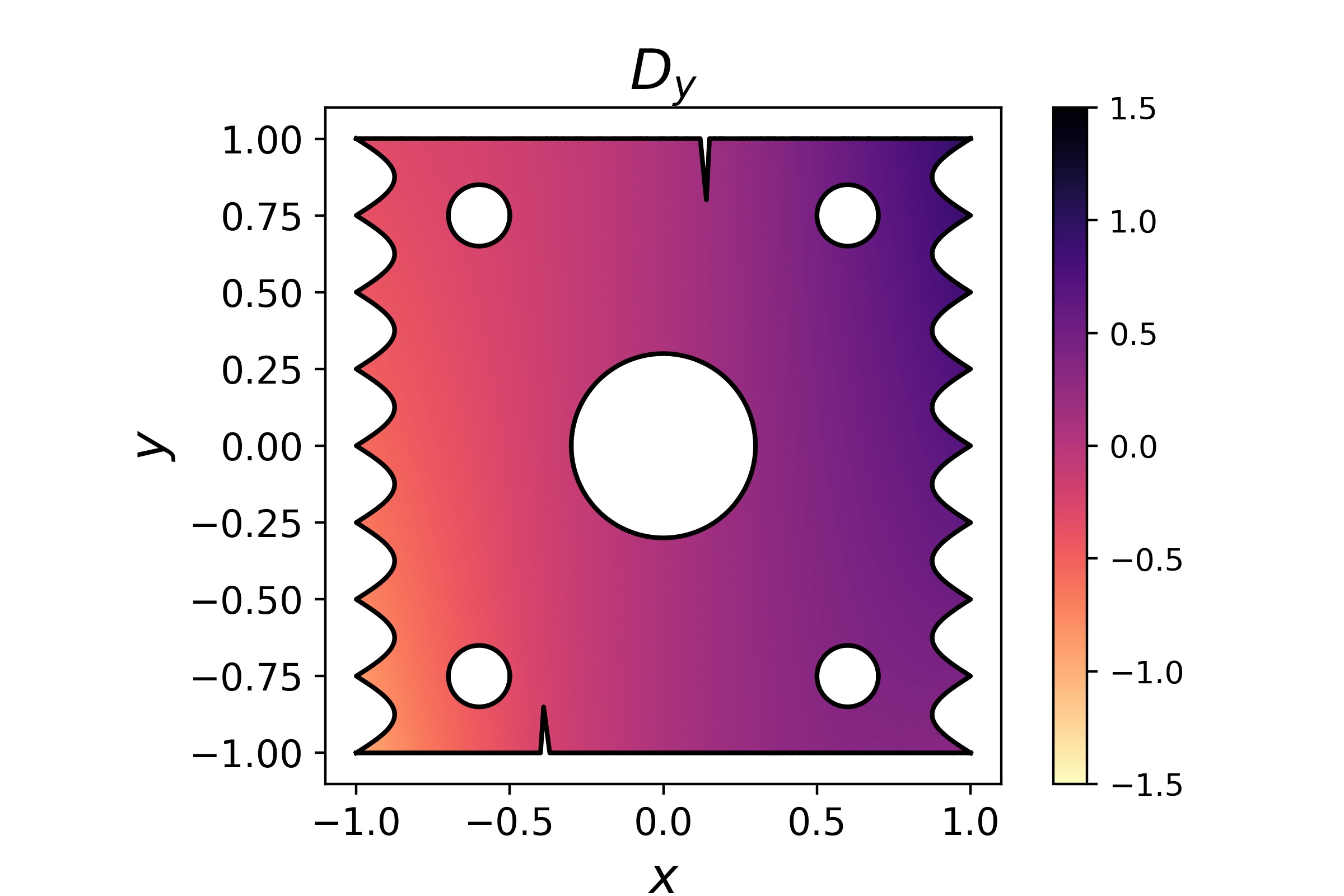}
			\caption{} 
		\end{subfigure}
		\captionsetup{}
	\end{center}
	\caption{Plate displacement: (a) x-axis displacement ($D_x$); (b) y-axis displacement ($D_y$).} 
	\label{E1_Disp}
\end{figure}

As a representative test case, we synthesize a governing equation that would generate a given solution.  In particular, we prescribe the plate displacement to take the following analytic form
\begin{align}\label{E1_analytic}
\begin{split}
u(x,y) & = \, 0.8 \sin \left( \frac{\pi}{2} (x +0.78) \right) \cos \left(y-1\right)   \\
& - 0.8  \sin \left( \frac{\pi}{2} (x +1.50) \right) \cos \left(y+1\right), \\
v(x,y) & = \, 0.72 - 0.65 \left[ \exp \left( \frac{-x^2y}{2} \right) +x \right].
\end{split} 
\end{align}
We further  consider an irregular  plate geometry that is shown in Figure \ref{E1_Disp} upon which  the two displacement fields of Equations~\ref{E1_analytic} are defined. With $E$ and $\nu$ set to 0.25 and 0.2, respectively, we can then analytically back-calculate the terms  $f_x$ and $f_y$ in Equation~\ref{E1_PDE} that would correspond to these prescribed geometry and displacement fields. Also, we establish a ``numerical" boundary condition $\mathcal{B}$, where for any given collocation point on the boundary, we set the displacement to be equal to the prescribed displacement fields $u_\mathcal{B}$ at those points.

To solve the PDE in Equation \ref{E1_PDE} using a PINN with the proposed PWC importance sampling approach, a neural network is constructed as the trial solution, with 4 hidden layers, each with 32 neurons. The input layer consists of two neurons, which take realizations from the physical domain (i.e., $x,y$). The output layer also consists of two neurons, which represent the horizontal and vertical plate displacement for the given realizations in the input layer. A Sine function is adopted for the activations in each hidden layer. The parameters of the trial solution are randomly initialized, and are trained following Algorithm \ref{Algorithm2}, with the following loss function

\begin{equation}
\resizebox{0.65\hsize}{!}{$
\begin{split}
& J  = J_1 + \lambda_2 J_2, \\
& J_1 = \bigg[ (\lambda+\mu) \frac{\partial}{\partial x} \left( \frac{\partial \tilde{u}}{\partial x} + \frac{\partial \tilde{v}}{\partial y} \right) +\mu \left( \frac{\partial^2 \tilde{u}}{\partial x^2} + \frac{\partial^2 \tilde{u}}{\partial y^2} \right) + f_x \\
 & +  (\lambda+\mu) \frac{\partial}{\partial y} \left( \frac{\partial \tilde{u}}{\partial x} + \frac{\partial \tilde{v}}{\partial y} \right) + \mu \left( \frac{\partial^2 \tilde{v}}{\partial x^2} + \frac{\partial^2 \tilde{v}}{\partial y^2} \right) + f_y \bigg]^2, \,\, (x,y) \in \mathcal{D}, \\ 
 & J_2 = \left( \tilde{u} - u_\mathcal{B} \right)^2,\,\,  (x,y) \in \partial \mathcal{D}.
\end{split}
$}
\end{equation}

The gradient of this loss function with respect to model parameters is computed through backpropagation \citep{baydin2018automatic}, as explained in Section \ref{DNN}. Parameter $\lambda_2$ set to 1 (note that parameter $\lambda_1$ is set to zero as the problem is time-independent). Batch size is set to 10,000, and the learning rate $\alpha$ is set to 0.002. A generalized Halton sequence generator algorithm is used to generate 100,000 collocation points and 10,000 seeds within the computation domain. Another 100,000 uniformly-distributed boundary collocation points are also generated. The model is trained for 500 iterations.

\begin{figure}
	\begin{center}
		\begin{subfigure}{0.48 \linewidth}
			\includegraphics[width=0.99\linewidth]{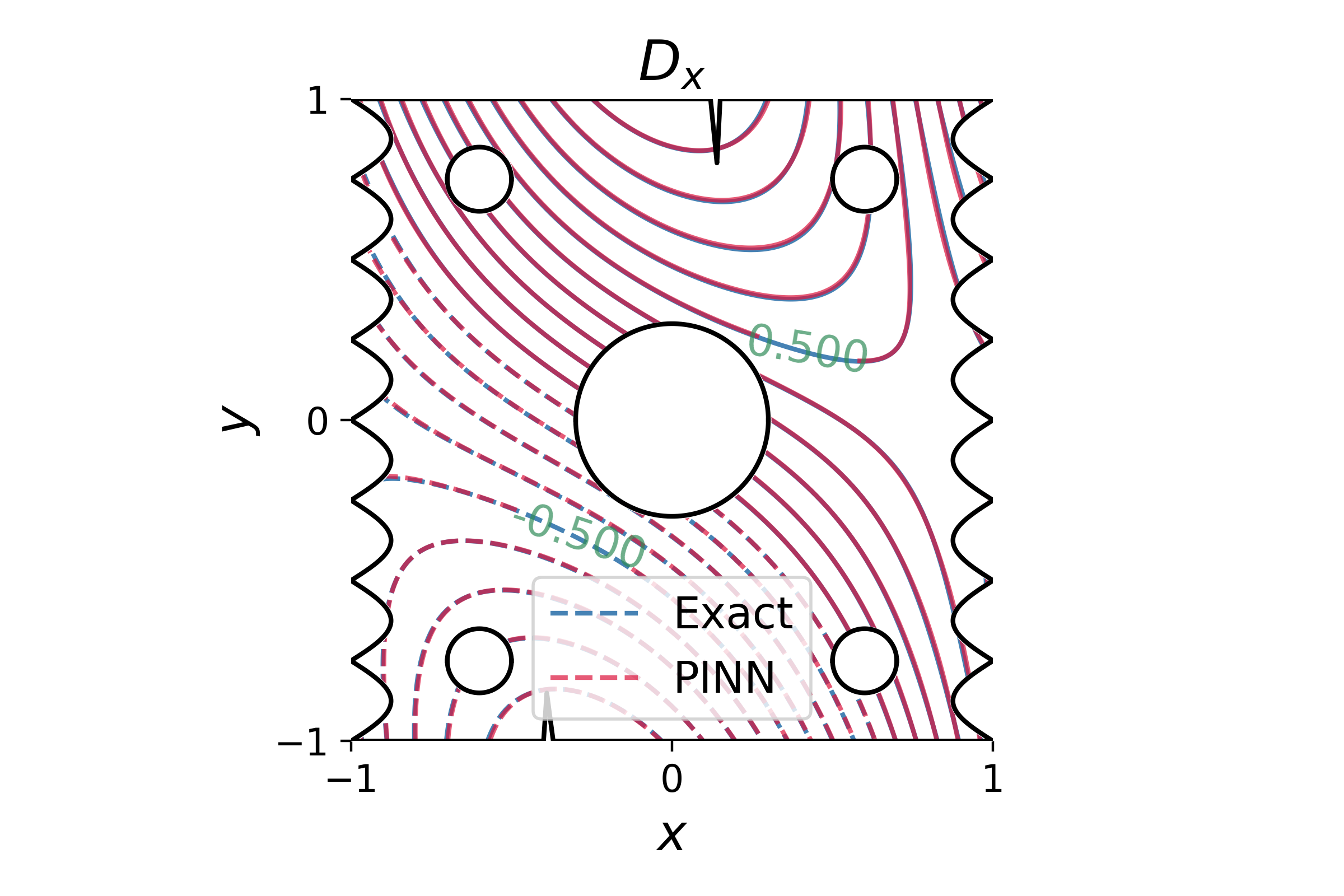}
			\caption{} 
		\end{subfigure}
		\quad
		\begin{subfigure}{0.48 \linewidth}
			\includegraphics[width=0.99\linewidth]{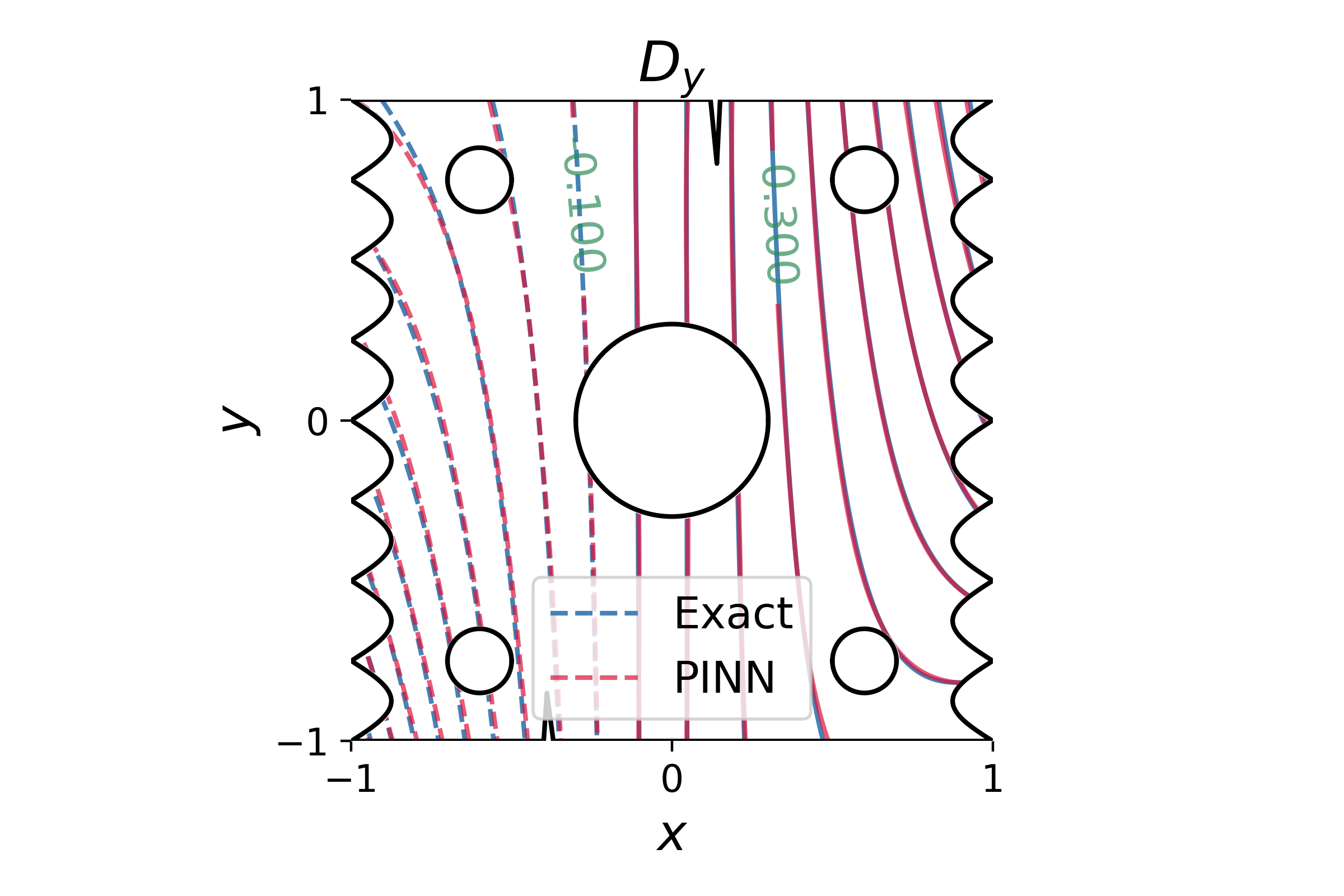}
			\caption{} 
		\end{subfigure}
		\captionsetup{}
	\end{center}
	\caption{A comparison between the exact and the PINN solutions to Equation \ref{E1_PDE}. The PINN solution is trained using the proposed importance sampling approach (Algorithm \ref{Algorithm2}) with piece-wise constant approximation to loss. $D_x$ and $D_y$ are, respectively, the displacement in $x$ and $y$ directions. The positive and negative values are shown by solid and dashed lines, respectively. } 
	\label{E1_contour}
\end{figure}

\begin{figure}
	\begin{center}
		\begin{subfigure}{0.7 \linewidth}
			\includegraphics[width=0.99\linewidth]{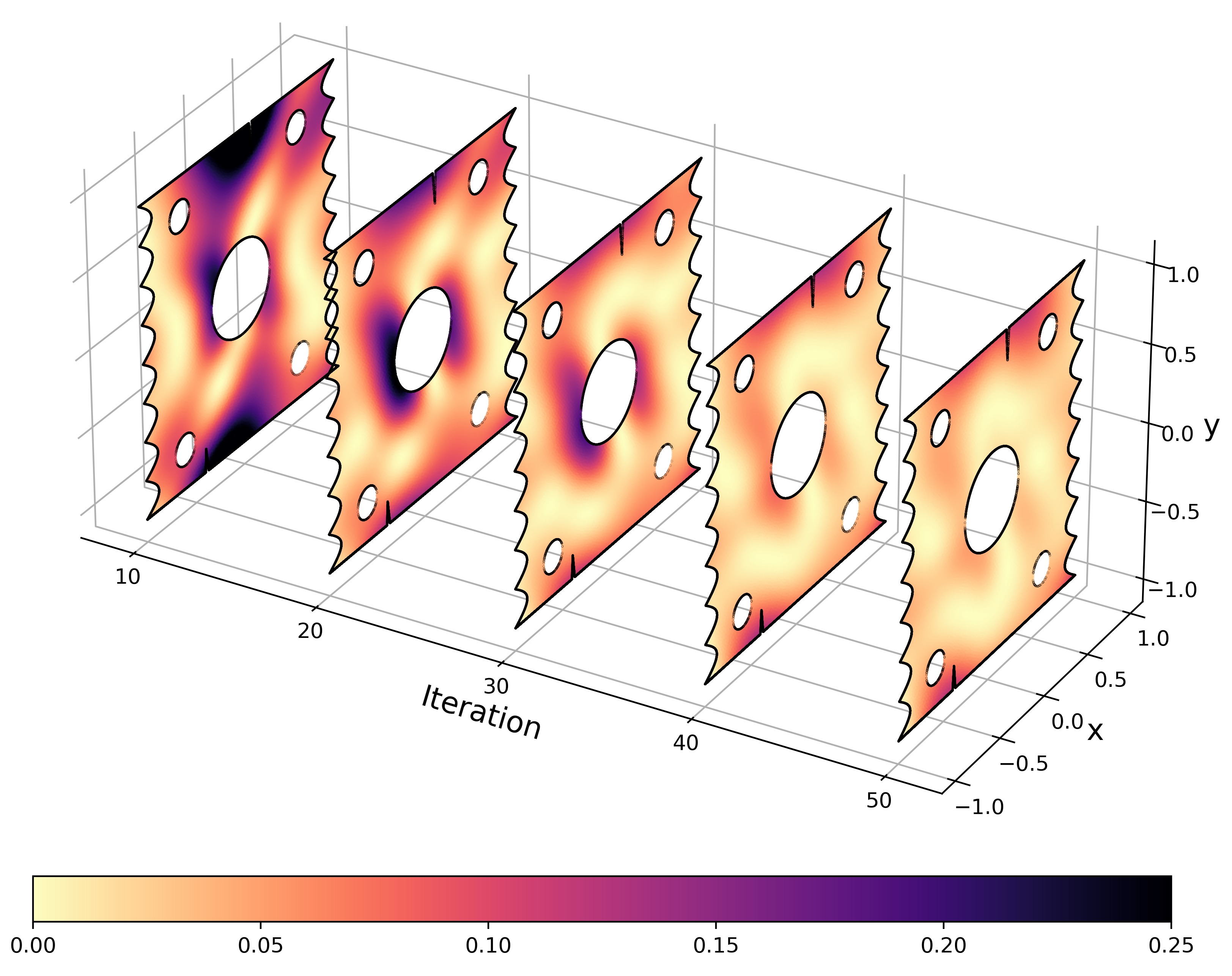}
			\caption{}
		\end{subfigure}
		\\
		\begin{subfigure}{0.7 \linewidth}
			\includegraphics[width=0.99\linewidth]{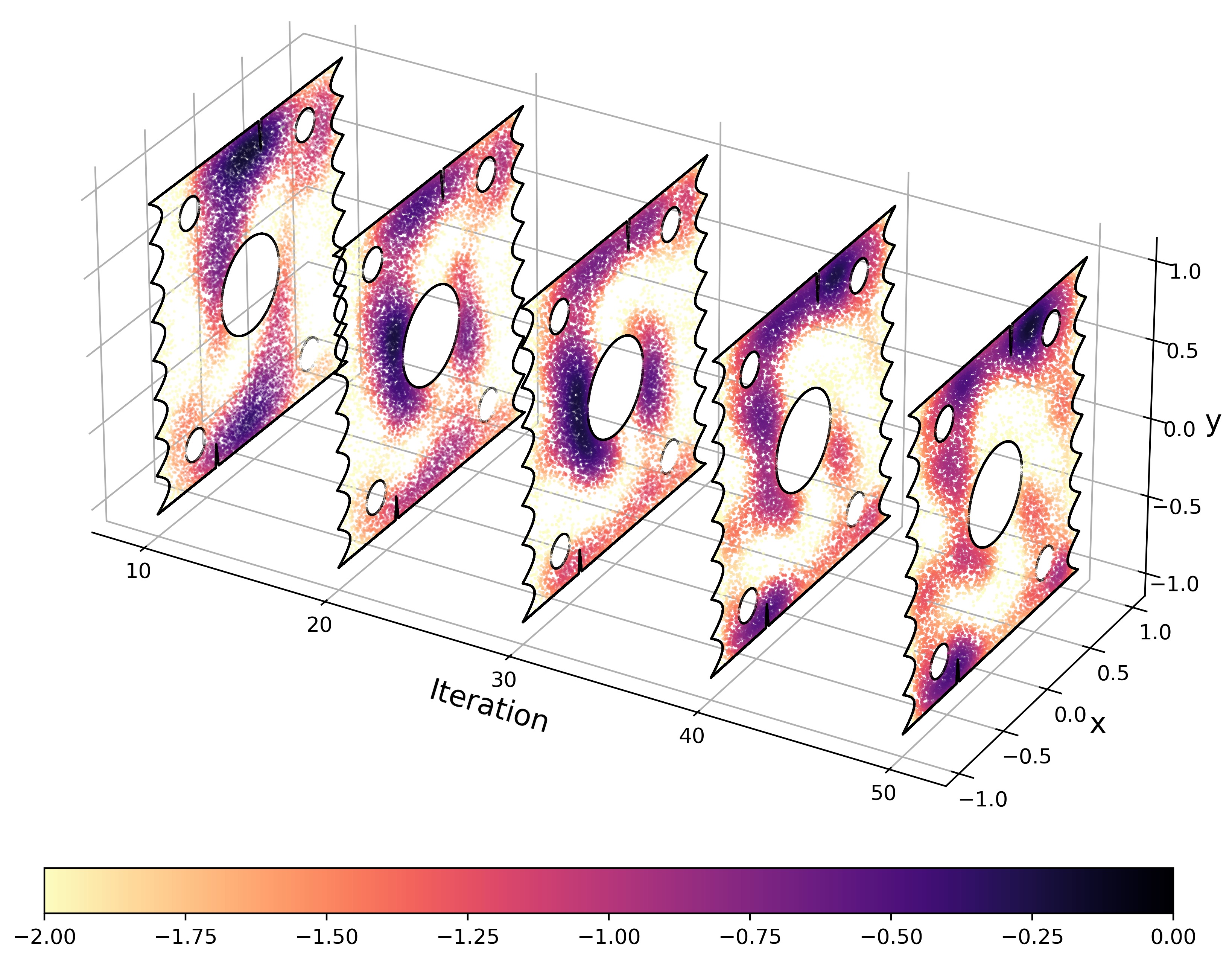}
			\caption{}
		\end{subfigure}
		\caption{a visualization of the progressive change in the (a) loss value, and (b) the sampled points, when the PINN solution to Equation \ref{E1_PDE} is trained using the proposed importance sampling approach. The color map in the two figures show, respectively, the loss value and the log probabilities showing the concentration of sampled points. } 
		\label{Levolution}
	\end{center}
\end{figure}

\begin{figure}[h]
	\begin{center}
		\begin{subfigure}{0.7 \linewidth}
			\includegraphics[width=0.99\linewidth]{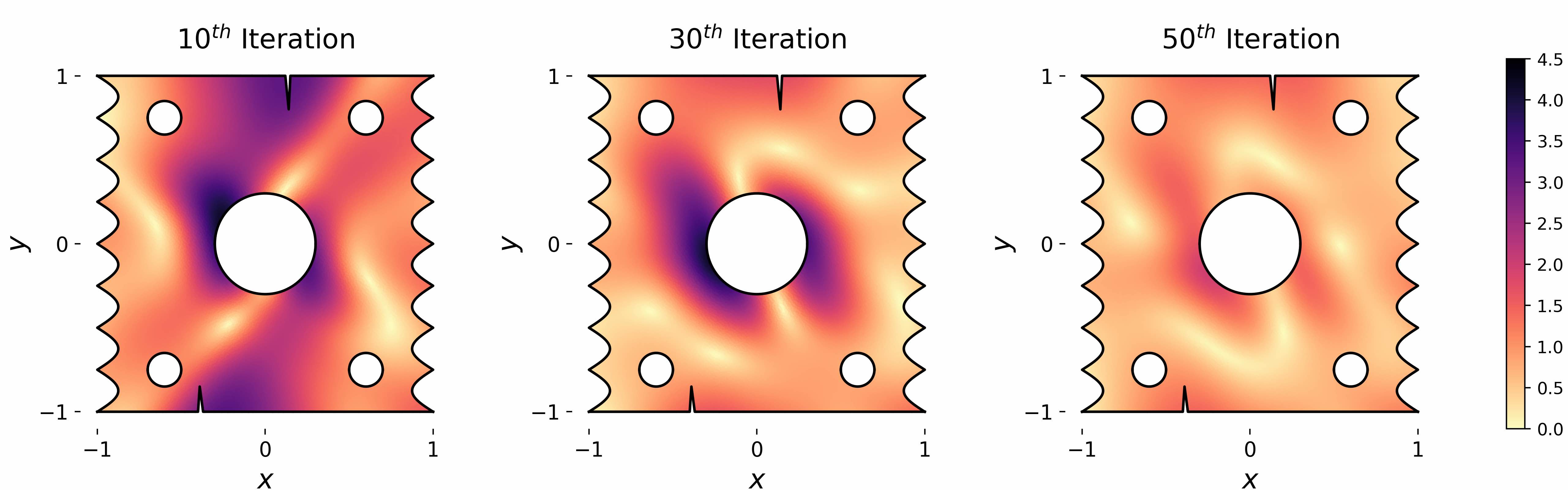}
			\caption{$\left \Vert \nabla_{\bm{\theta}}{J}(\bm{\theta}^{(i)};\bm x) \right \Vert _2$}
		\end{subfigure}
		\\
		\begin{subfigure}{0.7\linewidth}
			\includegraphics[width=0.99\linewidth]{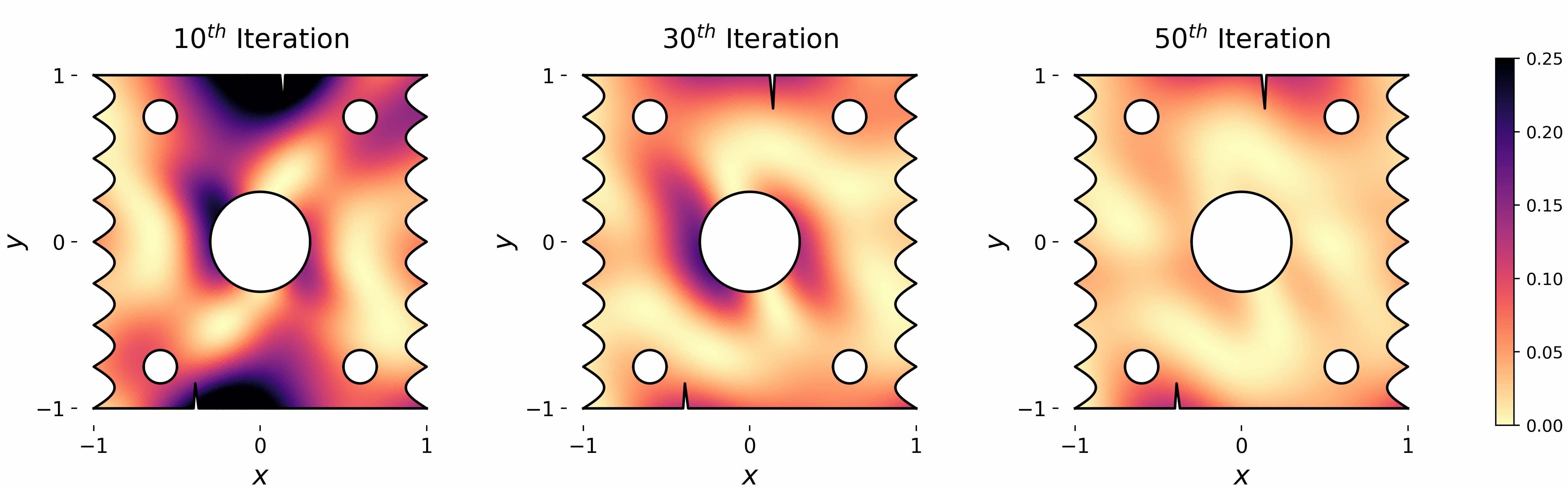}
			\caption{$  {J}(\bm{\theta}^{(i)};\bm x) $}
		\end{subfigure}
		\caption{Comparison between the 2-norm of the loss gradient w.r.t. model parameters (top) and the loss values (bottom) at three snapshots taken at iterations $i=10$, $i = 30$, and $i = 50$ of the proposed PWC importance sampling training. }
		\label{Gradient}
	\end{center}
\end{figure}

Figure \ref{E1_contour} presents a comparison between the PINN solution trained using the proposed approach and the exact solution in Equation \ref{E1_analytic}. It is evident that there is a close agreement between the results, verifying the accuracy of the proposed importance sampling approach. A visualization of the progressive change in the loss value as well as in the sampled points when the model is trained using the proposed approach is also presented in Figure \ref{Levolution}. Additionally, Figure \ref{Gradient} evaluates how well the gradient norm $\left \Vert \nabla_{\bm{\theta}}{J}(\bm{\theta}^{(i)};\bm x) \right \Vert _2$ used in Equation~\ref{sampling-gradient} is approximated by  $  {J}(\bm{\theta}^{(i)};\bm x)$ used in Equation~\ref{approx.q}. This comparison is shown for the loss fields at 3 different training steps of the  PWC importance sampling process.

\begin{figure}
	\begin{center}
		\begin{subfigure}{0.48 \linewidth}
			\includegraphics[width=0.99\linewidth]{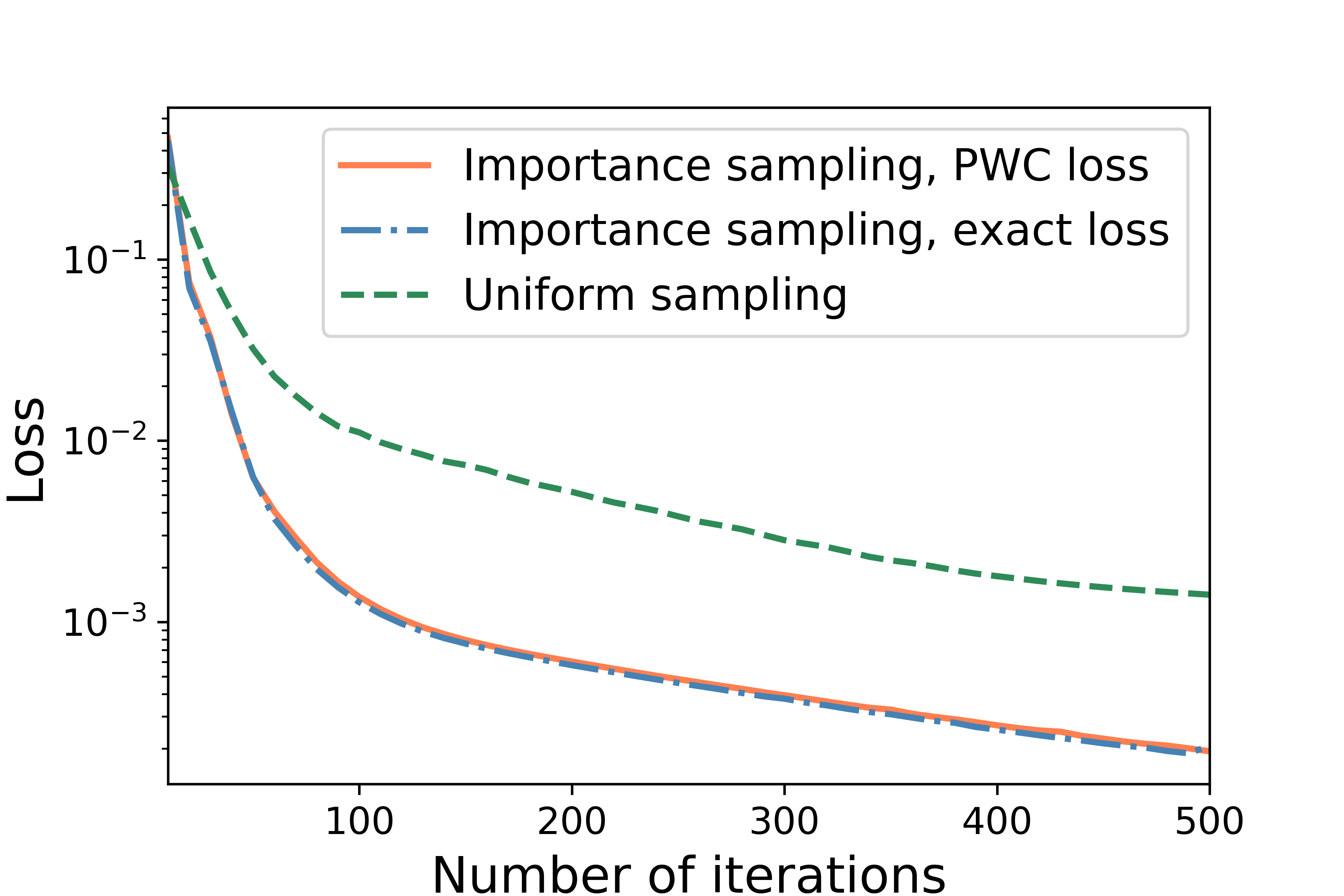}
			\caption{} 
			\label{E1_results.iter}
		\end{subfigure}
		\quad
		\begin{subfigure}{0.48 \linewidth}
			\includegraphics[width=0.99\linewidth]{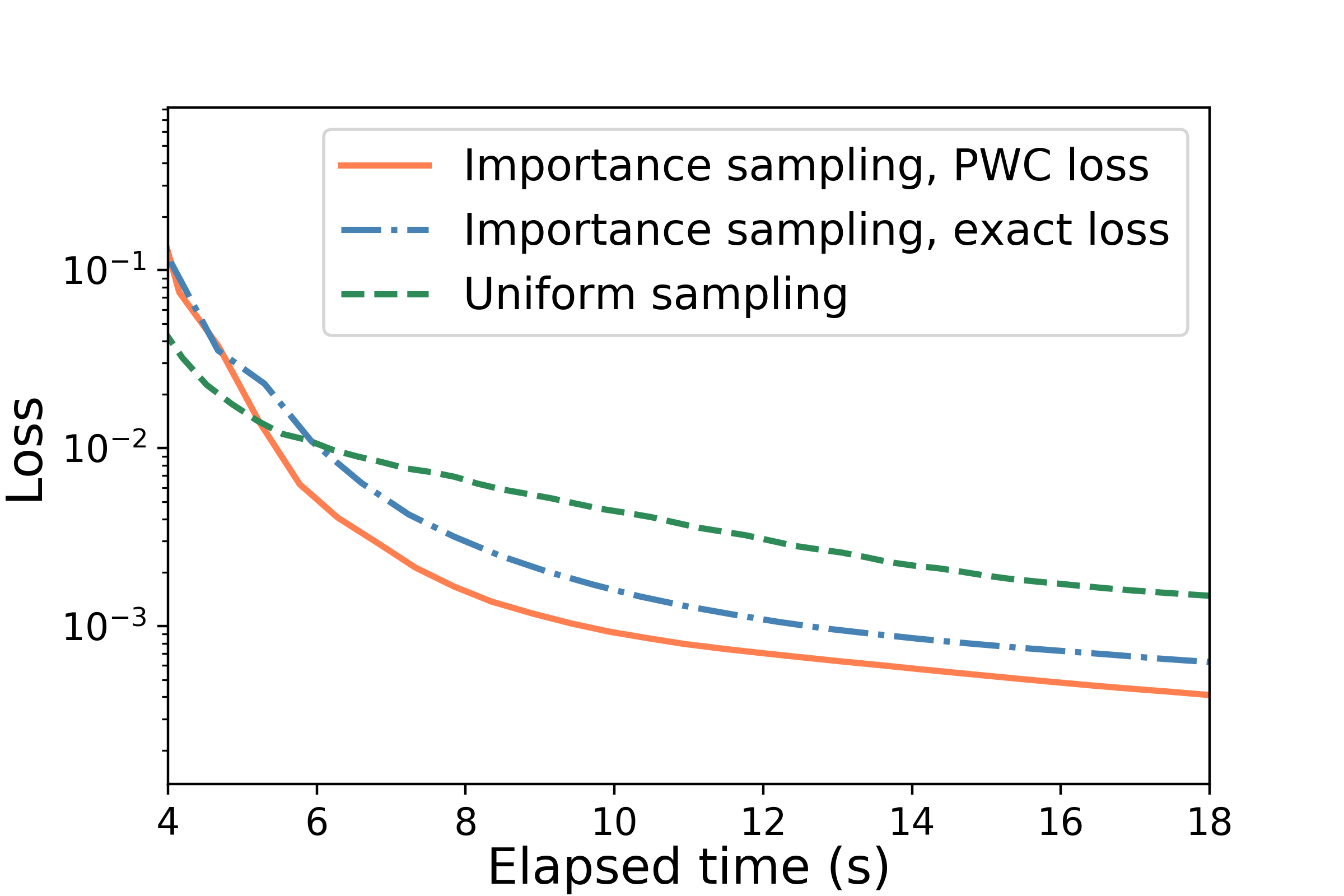}
			\caption{} 
			\label{E1_results.time}
		\end{subfigure}
		\captionsetup{}
	\end{center}
	\caption{A comparison between the performance of the three approaches of uniform sampling, importance sampling with exact loss evaluation, and importance sampling with approximate loss evaluation for training of a PINN solution to the elasticity problem defined in Equation \ref{E1_PDE}. } 
	\label{E1_results}
\end{figure}

To compare the accuracy and computational efficiency of the proposed PWC importance sampling approach with the uniform sampling approach, Figure \ref{E1_results} shows the loss value at different iterations (left) and different times (right) for each approach, and justifies the effectiveness of the PWC importance sampling method in accelerating the convergence of PINNs training. In comparing numerical performances, we also considered a third importance sampling approach which uses the exact loss values without any PWC approximation. It is evident from Figure~\ref{E1_results.iter} that the PWC approximation to loss is in fact a good approximation and does not negatively affect the convergence behavior. Figure~\ref{E1_results.time} shows that PWC importance sampling approach also outperforms the `exact loss' importance sampling method in terms of computational efficiency. This advantage is apparent in comparing computational times because the PWC importance sampling approach involves significantly less forward model evaluations, compared to the `exact loss' importance sampling method.

\begin{figure}[h]
	\begin{center}
		\includegraphics[width=0.5\linewidth]{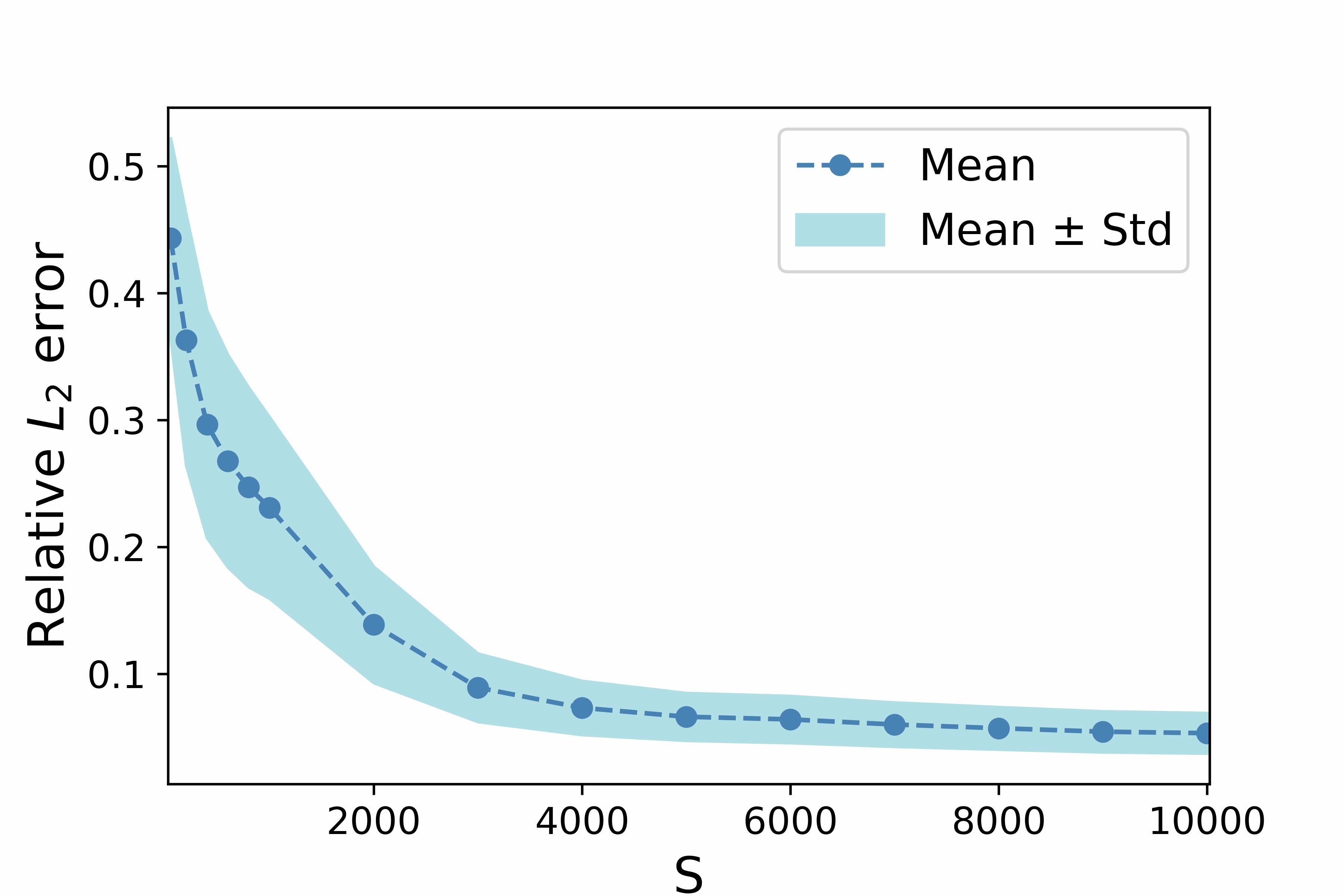}
		\caption{Mean and standard deviation, over the training iterations, of the relative $L_2$ error  in piece-wise constant approximation of loss values versus the number of seeds.}
		\label{Error}
	\end{center}
\end{figure}

\begin{figure}[h]
	\begin{center}
		\begin{subfigure}{0.48 \linewidth}
			\includegraphics[width=0.99\linewidth]{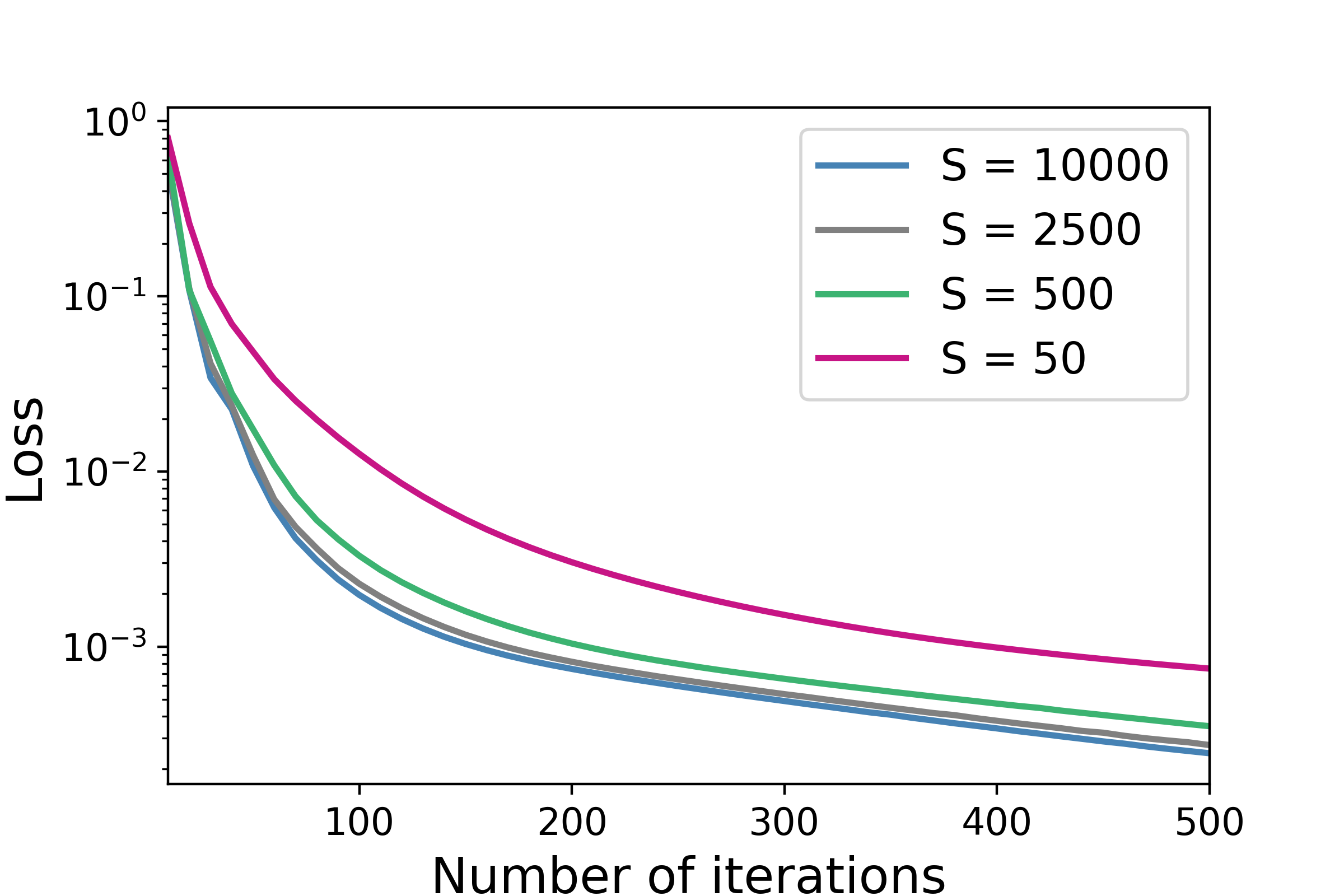}
			\caption{} 
		\end{subfigure}
		\quad
		\begin{subfigure}{0.48 \linewidth}
			\includegraphics[width=0.99\linewidth]{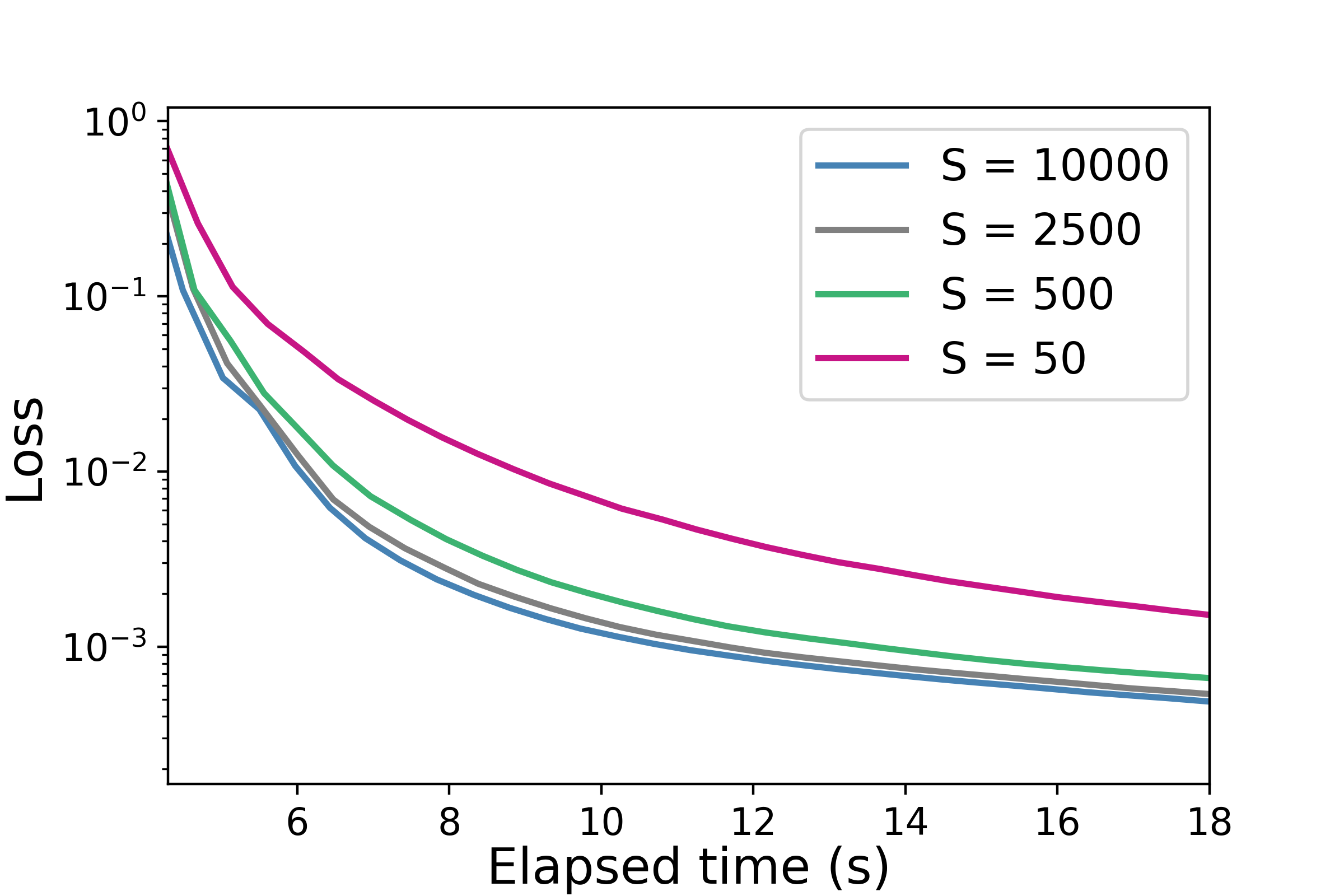}
			\caption{} 
		\end{subfigure}
		\captionsetup{}
	\end{center}
	\caption{A demonstration of the performance of the proposed PWC importance sampling approach with different choices for seed size, $S$. } 
	\label{50_seeds}
\end{figure}

\begin{figure}
	\begin{center}
		\begin{subfigure}{0.48 \linewidth}
			\includegraphics[width=0.99\linewidth]{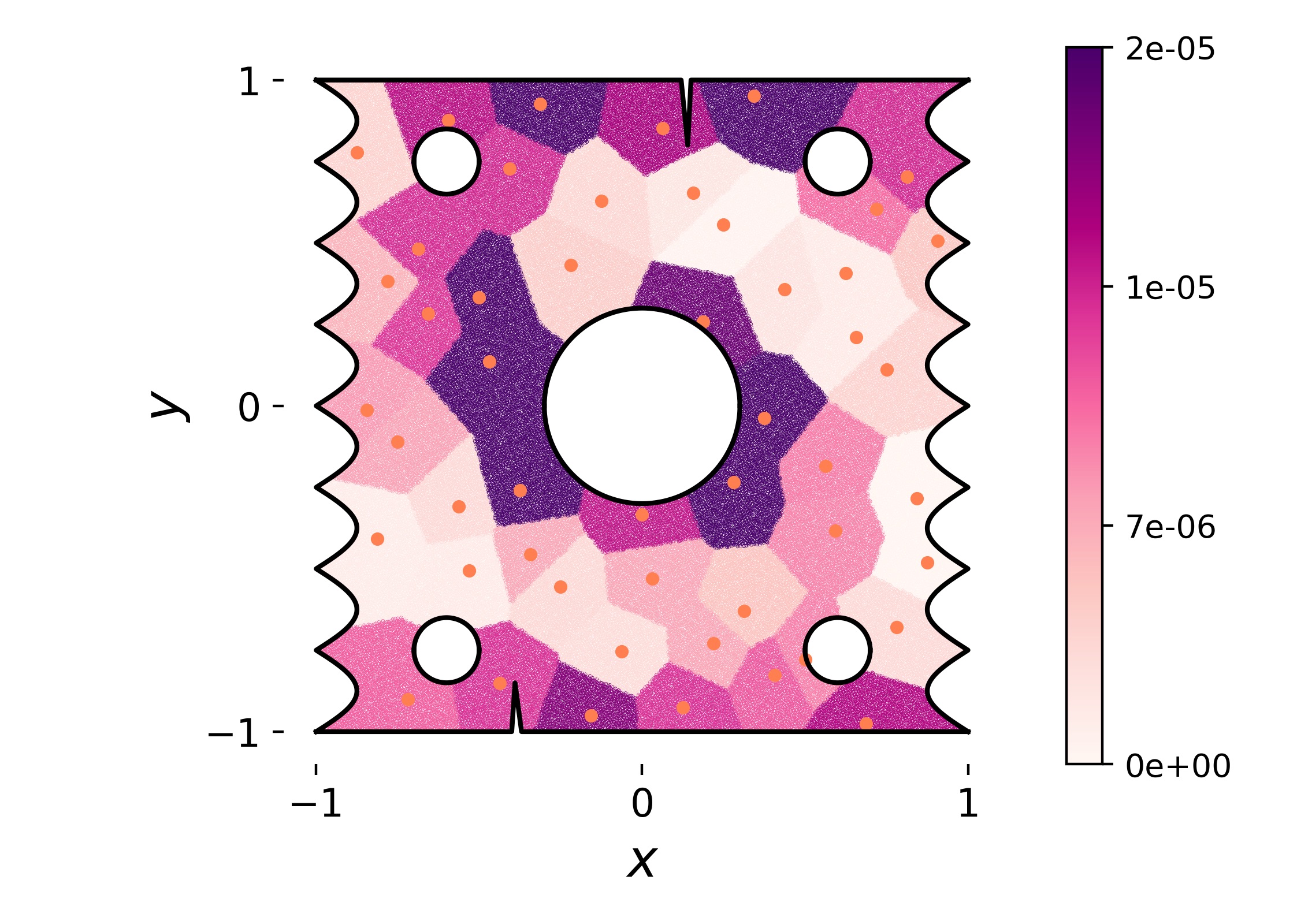}
			\caption{} 
		\end{subfigure}
		\quad
		\begin{subfigure}{0.48 \linewidth}
			\includegraphics[width=0.99\linewidth]{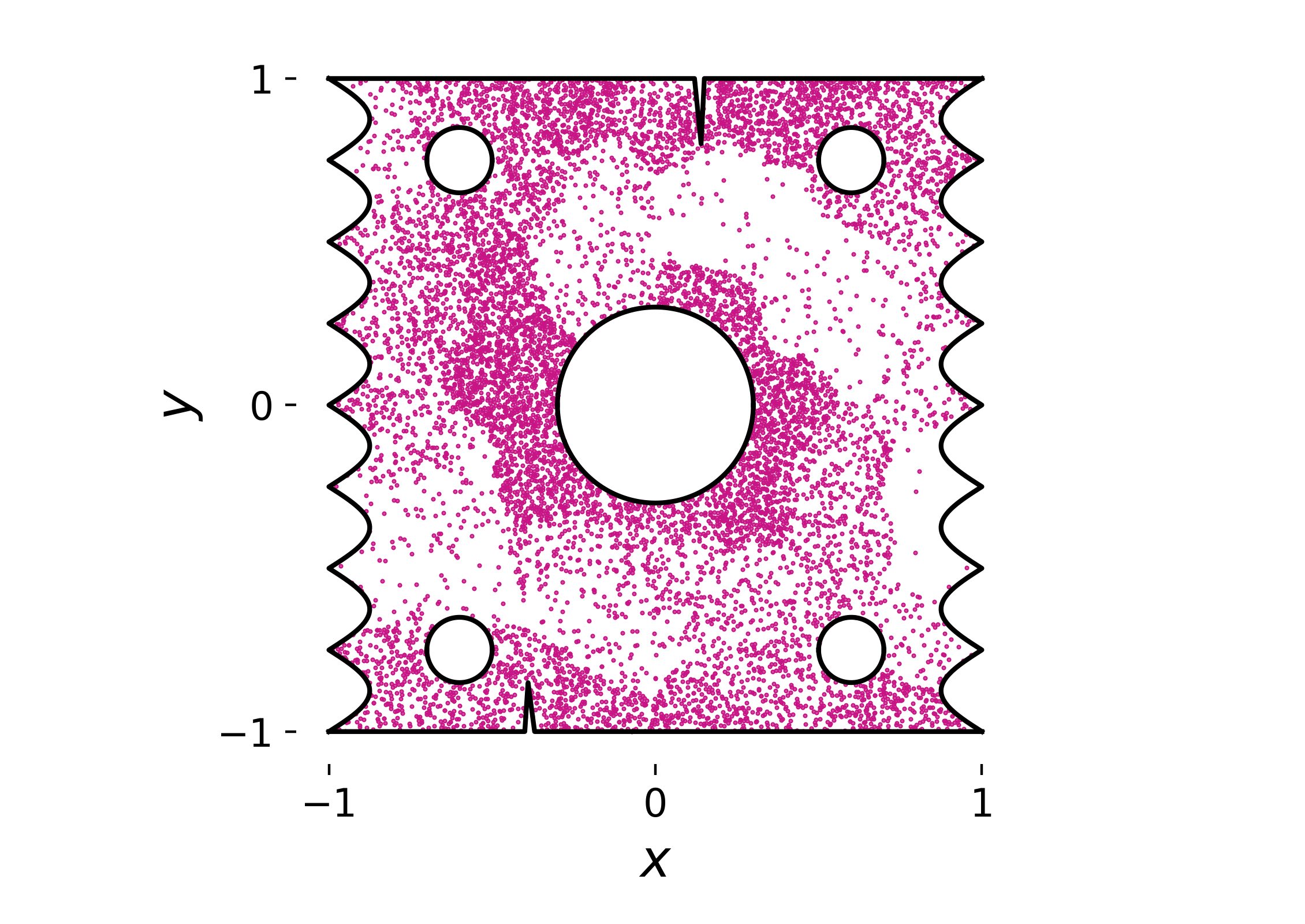}
			\caption{} 
		\end{subfigure}
		\captionsetup{}
	\end{center}
	\caption{(a) A visualization of the seeds and the collocation points color-coded with the selection probability at the $30^{th}$ training iteration of the proposed PWC importance sampling approach with $S=50$; (b) The 10,000 sampled collocation points. }
	\label{seeds.vis}
\end{figure}

Figure \ref{Error} shows how the error in piece-wise approximation of the loss function varies with respect to the number of seeds. In particular, for each seed size, the loss function approximation error is evaluated at and averaged over the $100,000$ collocation points and the variation in this averaged error over the 500 training epochs is depicted by the shaded area in this figure. Moreover, in order to demonstrate the effect of seed size selection on the convergence behavior of model training, Figure \ref{50_seeds} shows  loss values versus the number of iterations (left) and the elapsed time (right) for different  numbers of seeds. It is evident that the number of seeds, if selected reasonably (e.g. $S\ge500$ in this case), does not significantly affect the convergence behavior and therefore there is no need to consider that number as a new hyperparameter. This is because a PWC approximation with relatively large number of seeds (at least 500 in this case) can potentially provide a good approximation to the loss function, and increasing the number of steps may result in little or no change to the approximation accuracy. One reasonable suggestion is to simply set the seed size equal to the batch size. As a representative case for small seed sizes, we have also considered $S=50$, and included the corresponding performance in Figure~\ref{50_seeds}. For this seed size, Figure \ref{seeds.vis} shows the selection probabilities of collocation points, and also the 10,000 sampled collocation points, at the $30^{th}$ training iteration of the proposed PWC importance sampling approach. 

\subsection{\textcolor{black}{A plane stress problem}}\label{example3}

\begin{figure}[h]
	\begin{center}
		\includegraphics[width=0.4\linewidth]{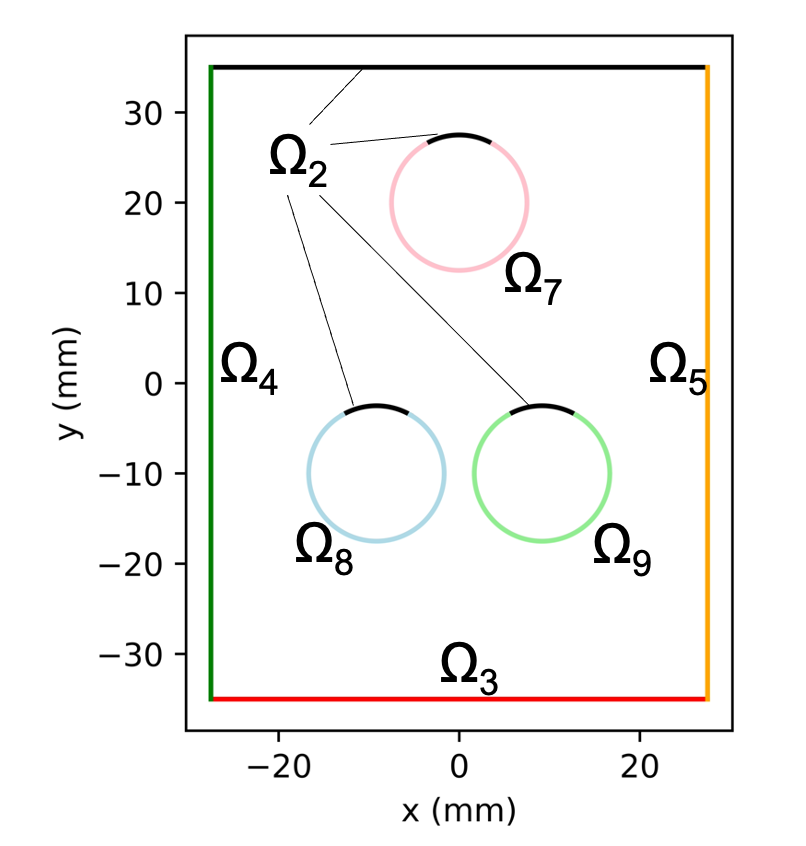}
		\caption{\textcolor{black}{The 2D plate with three holes considered for the solving linear elasticity problem using plane stress equations defined in Equation \ref{eqn:planestress}. The domains used in Equation \ref{eqn:losseqns} are labeled with different colors.} } 
		\label{E3_plate}
	\end{center}
\end{figure}

\textcolor{black}{In the second example, we demonstrate the performance of PINNs for a linear elasticity problem on a 2D plate with multiple holes using plane stress equations. In particular we focus on the mechanical behavior of cover-plates connections, which informs the  design of  joint and bolt positions  in the plates. This problem entails different behaviors such as the elastic-plastic behavior of the plate, the contact between bolts and holes, and the large displacements that could happen in particular configurations, which  collectively determine the failure mode of the plate, based on the complex stress distribution. It differs from the previous example in the following aspects. 
\begin{itemize}
    \item Here, there are stress boundary conditions and hence, Navier equations cannot be directly used.
    \item The mixed form of the equations used in this example are first order, whereas the Navier equations are second order. This can help increase the accuracy and the speed of the training.
    \item Since real material properties are used for the calculations, the values of these properties have to be normalized using an appropriate method.
\end{itemize}
For this experiment, following parameters used in the experimental work reported in \citep{TOUSSAINT2017148}, we consider a 6 mm thick plate of S235 grade steel with three holes for the bolts, two at the bottom and one at the top, as shown in Figure \ref{E3_plate}. A uni-axial tension is applied at the bottom of the plate in the form of an imposed displacement of 1.5 mm. The dimensions of plate are 55 mm $\times$ 70 mm. The holes have a radius of 7.5 mm. The coordinates of the centre of the three holes are $(0,20)$, $(-9.67,-10)$ and $(9.67,-10)$. The boundary conditions for the problem are 
\begin{itemize}
    \item $u_x=u_y=0$ for top edge of the plate (at $y=35.0$ mm) and for the top quarter of the three holes. Here, $u_x$ and $u_y$ are the displacements in the $x$ and $y$ directions respectively.
    \item $u_x=0,u_y=-1.5$ mm (imposed displacement) for the bottom edge of the plate (at $y=-35.0$ mm).
    \item Zero traction for all the free boundaries.
\end{itemize}}

\textcolor{black}{The governing plane stress equations for the problem are,}

\textcolor{black}{
\begin{equation}
\begin{split}\label{eqn:planestress}
&\hat{\sigma}_{xx} = \hat{\lambda} \left(\frac{\partial{\hat{u}}}{\partial{\hat{x}}} + \frac{\partial{\hat{v}}}{\partial{\hat{y}}} + \frac{-\hat{\lambda}}{\hat{\lambda}+2\hat{\mu}}(\frac{\partial{\hat{u}}}{\partial{\hat{x}}} + \frac{\partial{\hat{v}}}{\partial{\hat{y}}})\right)+2\hat{\mu} \frac{\partial{\hat{u}}}{\partial{\hat{x}}}.\\
&\hat{\sigma}_{yy} = \hat{\lambda} \left(\frac{\partial{\hat{u}}}{\partial{\hat{x}}} + \frac{\partial{\hat{v}}}{\partial{\hat{y}}} + \frac{-\hat{\lambda}}{\hat{\lambda}+2\hat{\mu}}(\frac{\partial{\hat{u}}}{\partial{\hat{x}}} + \frac{\partial{\hat{v}}}{\partial{\hat{y}}})\right)+2\hat{\mu} \frac{\partial{\hat{v}}}{\partial{\hat{x}}}.
\end{split}
\end{equation}
where $\hat{u},\hat{v}$ are the non-dimensionalized, normalized displacement vectors, $\hat{x}$ and $\hat{y}$ are the normalized directions. $\hat{\lambda}$ and $\hat{\mu}$ are the corresponding non-dimensionalized, normalized Lam\'e constants. These values are given by $\hat{x_i}={x_i}/{L}, \hat{u_i}={u_i}/{U}, \hat{\lambda}={\lambda}/{\mu_c}, \hat{\mu}={\mu}/{\mu_c}$, where, $L$ is the characteristic length, $U$ is the characteristic displacement, and $\mu_c$ is the non-dimensionalizing shear modulus. The non-dimensionalized form of the stress-displacement equations are given by \begin{equation}
\begin{split}\label{eqn:stressdisp}
&\hat{\sigma}_{ij}=\hat{\lambda}\hat{\epsilon}_{kk}\delta_{ij} + 2\hat{\mu}\hat{\epsilon}_{ij}\\
\end{split}
\end{equation} where, $\delta_{ij}$ is the Kronecker delta function and $\hat{\epsilon}_{ij}$ is the strain tensor that takes the following form
\begin{equation}
\begin{split}\label{eqn:straineqn}
&\hat{\epsilon}_{ij}=\frac{1}{2}(\hat{u}_{i,j}+\hat{u}_{j,i}).\\
\end{split}
\end{equation}The non-dimensionalized plane stress equations are adopted from SimNet (\cite{hennigh2020nvidia,simnet2020}).}

\textcolor{black}{To solve this PDE using a PINN, we constructed a neural network with 4 hidden layers with 32 units in each hidden layer. The input layer consists of two neurons, which take realizations from the physical domain, $\hat{x}$ and $\hat{y}$ . The output layer consists of five neurons representing the solution ($\tilde{u},\tilde{v}$,$\tilde{\sigma}_{xx}$, $\tilde{\sigma}_{xy}$, $\tilde{\sigma}_{yy}$) for the given realizations in the input layer. We adopt {a Swish function} for the activation in each hidden layer. The parameters of the neural network are randomly initialized, and are trained following Algorithm 2, with the following loss function (domains used in the equations below are shown in Figure \ref{E3_plate}.)}

\textcolor{black}{
\begin{equation}
	\begin{split}\label{eqn:losseqns}
	  & \quad \quad \quad \quad \quad \quad J  = \sum_{i=1}^9 \lambda_i  J_i,  \\
	 J_1 & = \left(\frac{\partial{\tilde{\sigma}_{xx}}}{\partial{\hat{x}}}+\frac{\partial{\tilde{\sigma}_{xy}}}{\partial{\hat{y}}}\right)^2 + \left(\frac{\partial{\tilde{\sigma}_{xy}}}{\partial{\hat{x}}}+\frac{\partial{\tilde{\sigma}_{yy}}}{\partial{\hat{y}}}\right)^2, \\ 
	 J_2 & = (\tilde{u}_x-0.0)^2+(\tilde{u}_y-0.0)^2,  \quad \quad \forall x, y \in \Omega_2, \\
	 J_3 & = (\tilde{u}_x-0.0)^2+(\tilde{u}_y+1.0)^2,  \quad \quad \forall x, y \in \Omega_3, \\
	 J_4 & = \tilde{\sigma}_{xx}^2 + \tilde{\sigma}_{xy}^2,\quad \quad \quad \quad \quad \quad \quad \quad \forall x, y \in \Omega_4, \\
	 J_5 & = \tilde{\sigma}_{xx}^2 + \tilde{\sigma}_{xy}^2,\quad \quad \quad \quad  \quad \quad \quad\quad \forall x, y \in \Omega_5, \\
	 J_6 & = ( \hat{\sigma}_{xx}- \tilde{\sigma}_{xx})^2 + ( \hat{\sigma}_{xy}- \tilde{\sigma}_{xy})^2 + ( \hat{\sigma}_{yy}- \tilde{\sigma}_{yy})^2,\\
	 J_7 &= ((\tilde{\sigma}_{xx}  x+\tilde{\sigma}_{xy}  (y-20.0))/7.5)^2,  \forall x, y \in \Omega_7, \\
	 J_8 &= ((\tilde{\sigma}_{xx}  (x+9.17)+\tilde{\sigma}_{xy}  (y+10.0))/7.5)^2, \forall x, y \in \Omega_8, \\
	 J_9 & = ((\tilde{\sigma}_{xx} (x-9.17)+\tilde{\sigma}_{xy}  (y+10.0))/7.5)^2, \forall x, y \in \Omega_9.\\
	\end{split}
\end{equation}}

\begin{figure}
	\begin{center}
		\begin{subfigure}{0.48 \linewidth}
			\includegraphics[width=0.99\linewidth]{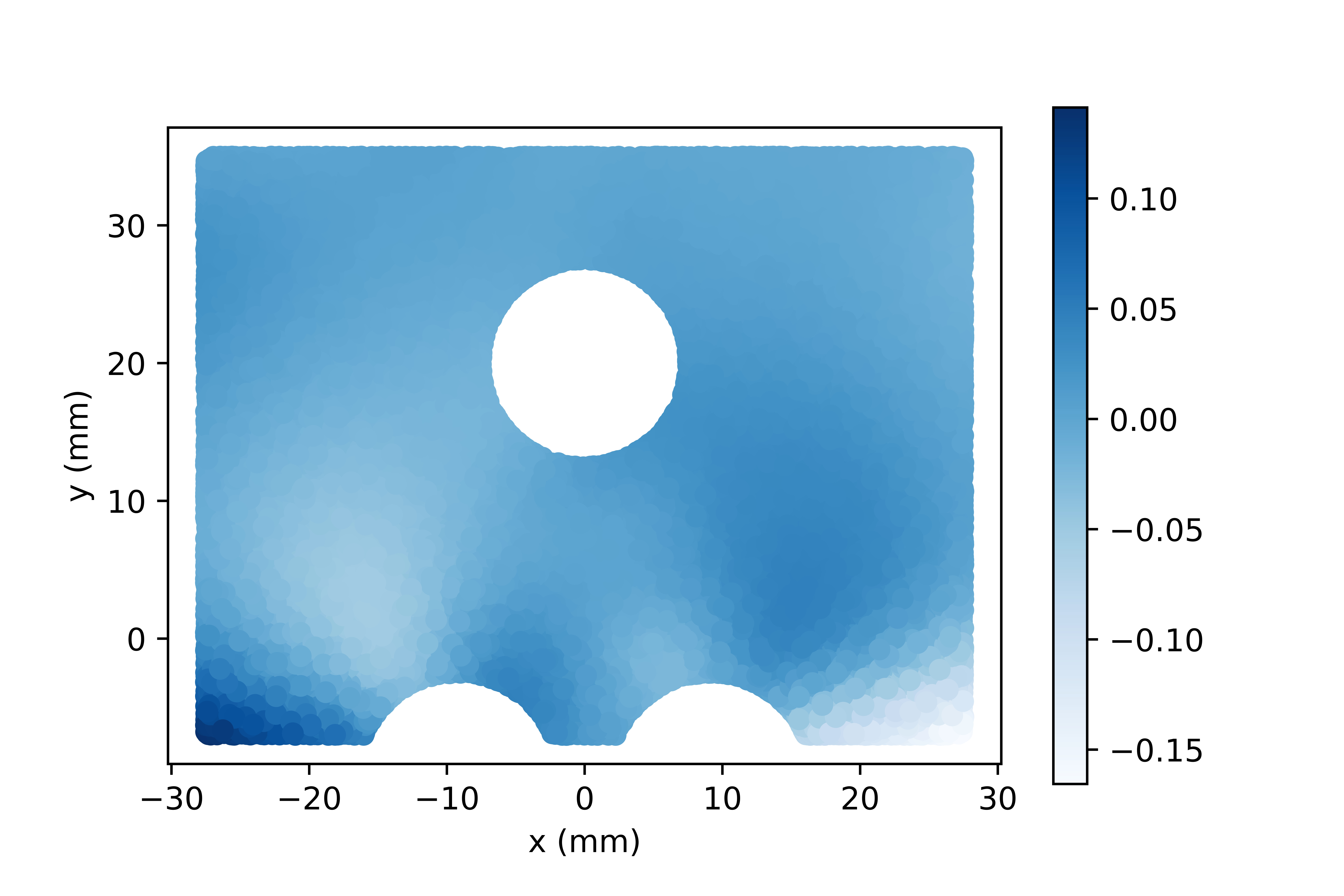}
			\caption{Displacement in the x-direction} 
		\end{subfigure}
		\quad
		\begin{subfigure}{0.48 \linewidth}
			\includegraphics[width=0.99\linewidth]{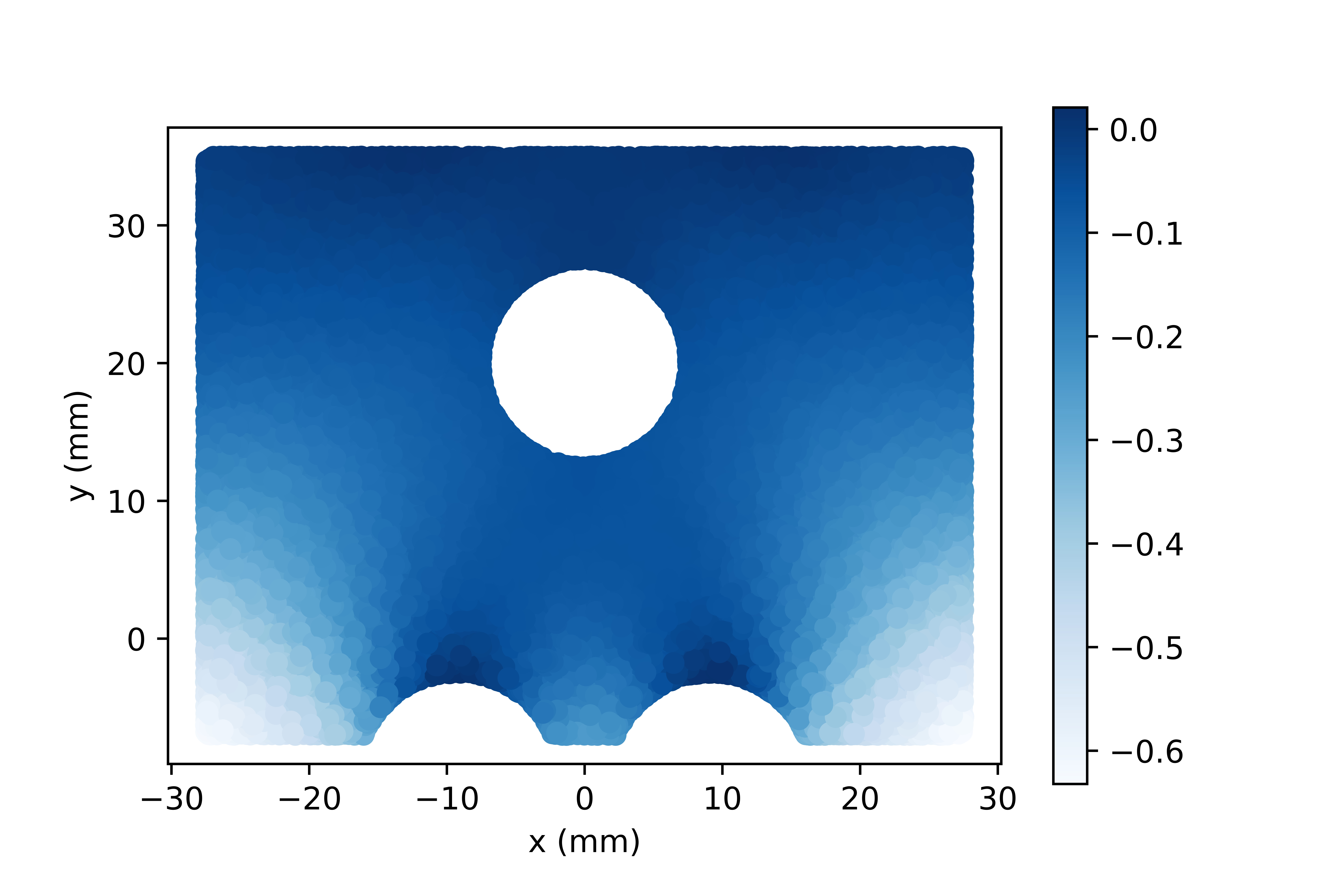}
			\caption{Displacement in the y-direction} 
		\end{subfigure}
		\captionsetup{}
	\end{center}
	\caption{\textcolor{black}{Plane displacement computed by the PINN model using importance sampling.}}
	\label{E3.u}
\end{figure}

\begin{figure}[h]
	\begin{center}
		\begin{subfigure}{0.48 \linewidth}
			\includegraphics[width=0.99\linewidth]{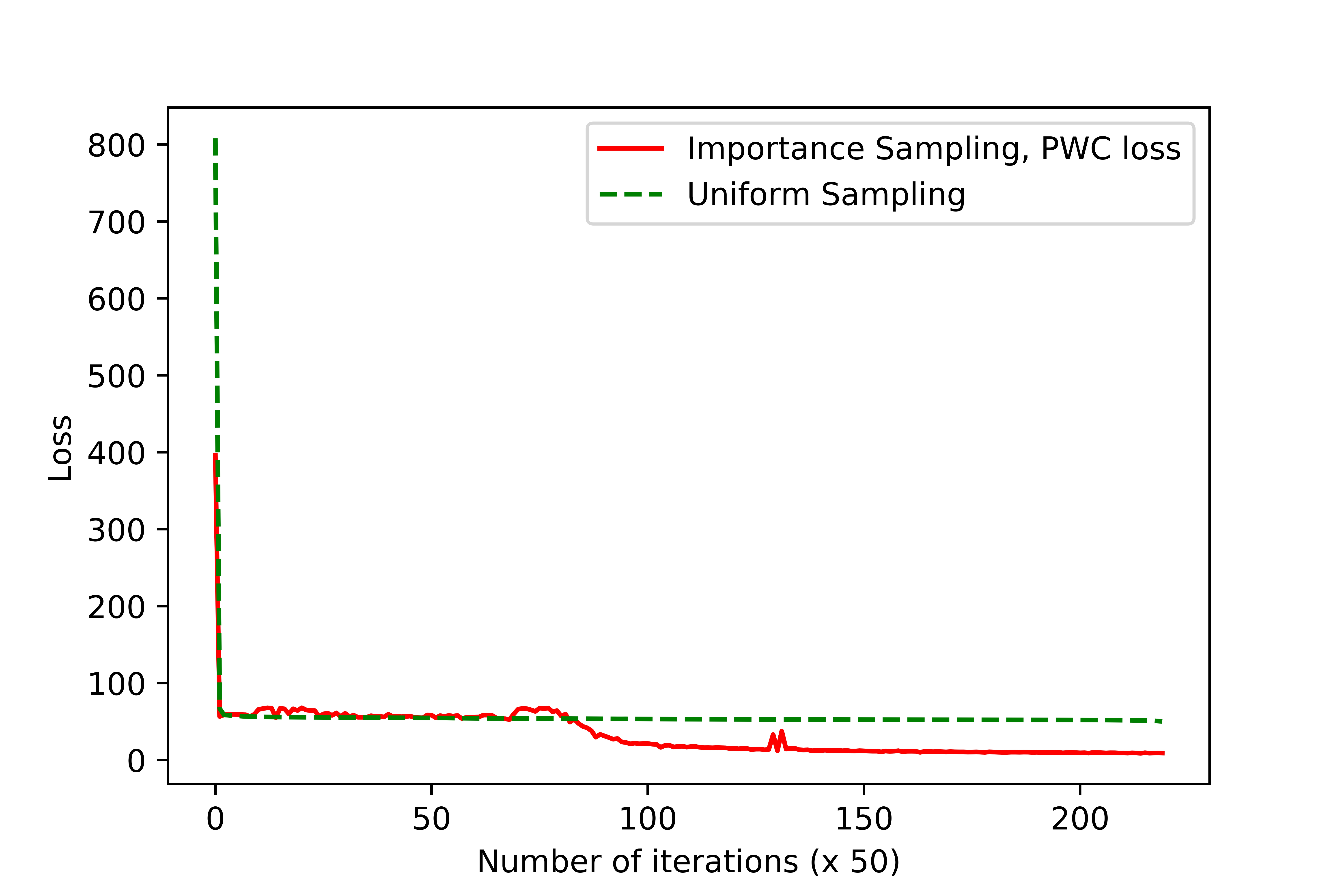}
			\caption{} 
			\label{E3_results.iter}
		\end{subfigure}
		\quad
		\begin{subfigure}{0.48 \linewidth}
			\includegraphics[width=0.99\linewidth]{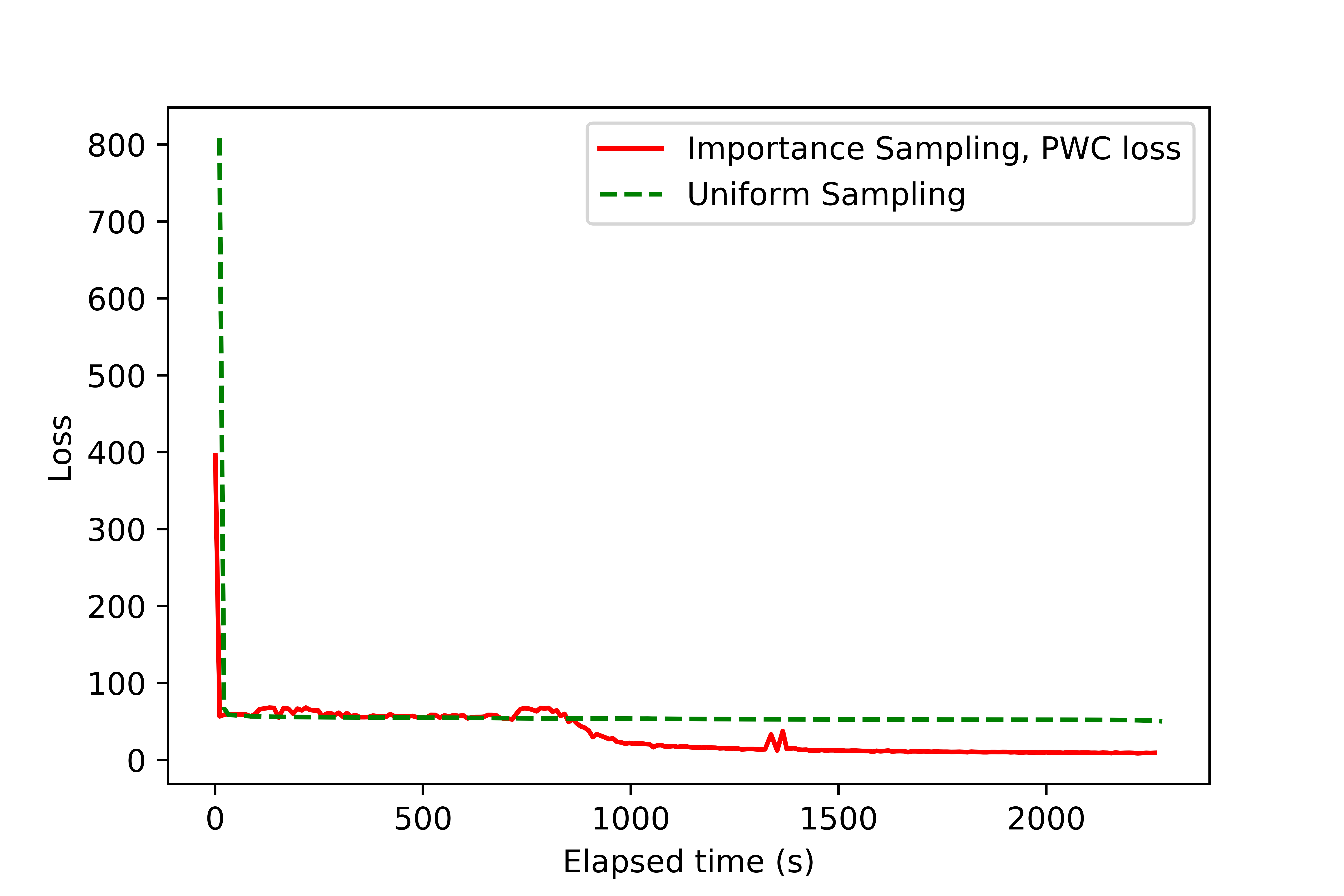}
			\caption{} 
			\label{E3_results.time}
		\end{subfigure}
		\captionsetup{}
	\end{center}
	\caption{\textcolor{black}{Comparisons between the training loss of PINN with importance sampling and PINN with uniform sampling  when applied on the plane stress of Section \ref{example3}. Training losses are depicted versus the number of iterations (top figure), and elapsed time (bottom figure).}}
	\label{E3_results}
\end{figure}

\textcolor{black}{The gradient of the loss function with respect to model parameters are computed through backpropagation and parameters $\lambda_1$ through $\lambda_9$, which determine the contributions of each loss value to the overall loss, are finalized through some iterations of training with few epochs. The final values of these parameters are set to be $\lambda_1=500$, $\lambda_2=1000$, $\lambda_3=1000$, $\lambda_4=75$, $\lambda_5=75$, $\lambda_6=200$, $\lambda_7=75$, $\lambda_8=75$, $\lambda_9=75$. Batch size is set to be 5000 and the learning rate is 0.0005. A generalized Halton sequence generator algorithm is used to generate N=500,000 collocation points and S=5,000 seeds within the computation domain. Another 500,000 uniformly-distributed boundary collocation points are also generated. The model is trained for 25,000 iterations. Figure \ref{E3.u} shows the computed displacements obtained by our proposed importance sampling approach. Furthermore, Figure \ref{E3_results} shows that in training the PINN model, the importance sampling approach provides better computational performance compared to the uniform sampling approach, in terms of both the number of iterations and elapsed time.}

\subsection{A transient diffusion problem}\label{example2}

\begin{figure}
	\begin{center}
		\includegraphics[width=0.6\linewidth]{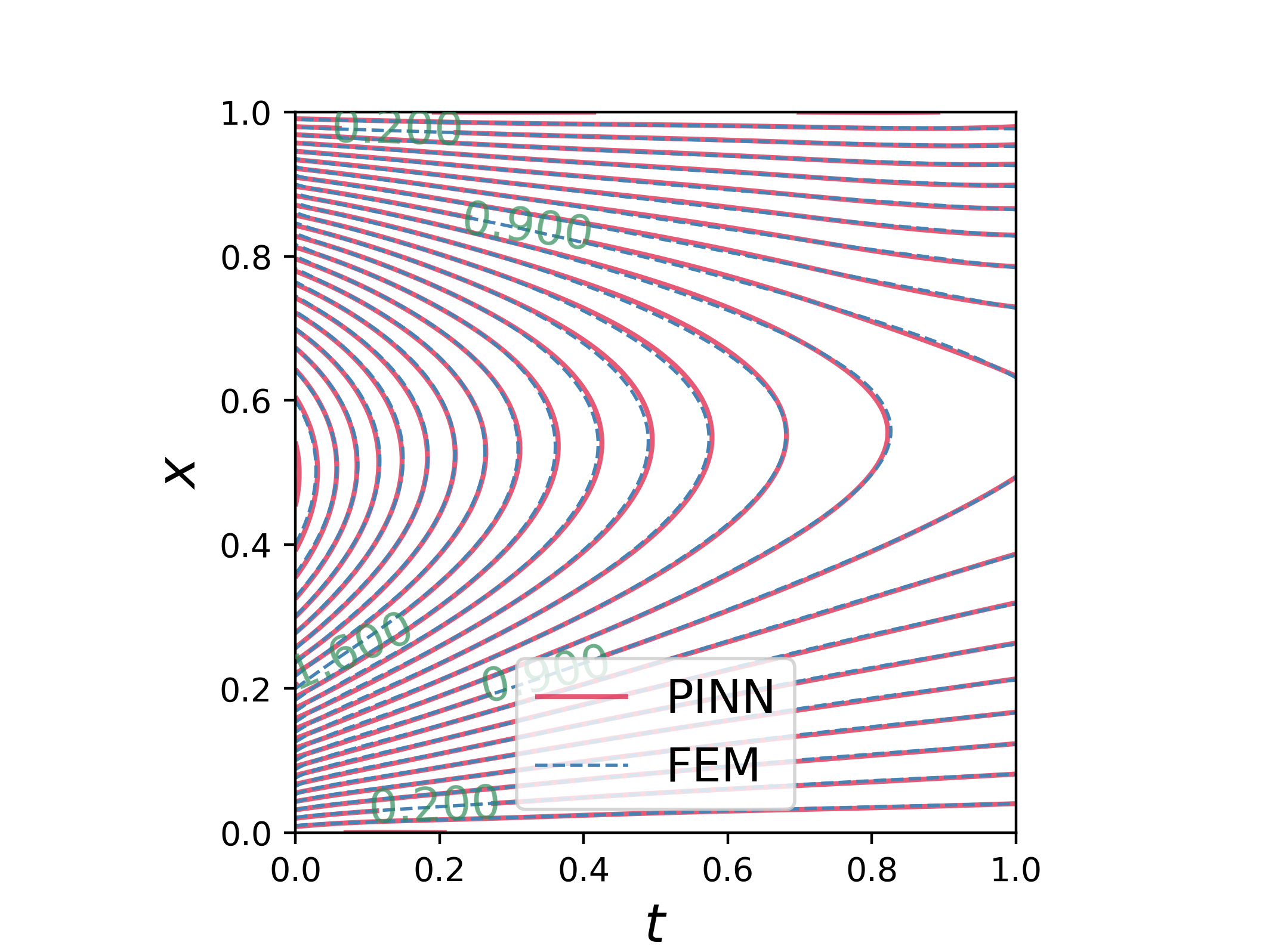}
		\caption{A comparison between the Finite Element and the PINN solutions for $u$ in Equation \ref{eqn:diffusion}. The PINN solution is trained using the proposed importance sampling approach with piece-wise constant approximation to loss.} 
		\label{E2_contour}
	\end{center}
\end{figure}

In the previous two examples, we demonstrated the performance of our proposed PWC importance sampling approach on a boundary value problem. However, in addition to boundary value problems, PINNs have  also been applied  to solve time-dependent PDEs (e.g. \cite{raissi2019physics,nabian2020physics}). Therefore, to numerically verify the accuracy and efficiency of our proposed method in solving time-dependent problems, let us consider a transient diffusion problem, governed  by the following PDE,

\begin{equation}
\begin{split}\label{eqn:diffusion}
&\mathcal{N}(t,x,u) =  \frac{\partial{u}}{\partial{t}} - \frac{\partial^2{u}}{\partial{x^2}} - 3x,\; \; \;  t\in\left [ 0,1 \right ], x\in\left [ 0,1 \right ], \\
&\mathcal{I}(x) =  u -  10(x-x^2), \; \; \;  x\in\left [ 0,1 \right ], \\
&\mathcal{B}(t,x) =  u, \; \; \;  t\in\left [ 0,1 \right ],x\in \{0,1 \}.
\end{split}
\end{equation}
To solve this PDE using a PINN we construct as the trial solution a neural network with 4 hidden layers with 32 units in each hidden layer. The input layer consists of two neurons, which take realizations from the physical domain (i.e., $t,x$). The output layer consists of one neuron, which represents the solution $u$ for the given realizations in the input layer. We adopt a Sine function for the activations in each hidden layer. The parameters of the trial solution are randomly initialized, and are trained following Algorithm \ref{Algorithm2}, with the following loss function
\begin{equation}
	\begin{split}
	& J  = J_1 + \lambda_1 J_2 + \lambda_2 J_3,  \\
	& J_1 = \left( \frac{\partial{\tilde{u}}}{\partial{t}} - \frac{\partial^2{\tilde{u}}}{\partial{x^2}} - 3x \right)^2, \,\, t \in [0,1], x \in [0,1], \\ 
	& J_2 = \left( \tilde{u} -  10(x-x^2) \right)^2,\,\,  t=0, x \in [0,1], \\ 
	& J_3 =  \tilde{u}^2, \,\,\,  t \in [0,1], x \in \{0,1\}. 
	\end{split}
\end{equation}
The gradient of this loss function with respect to model parameters is computed through backpropagation \citep{baydin2018automatic}, as explained in Section \ref{DNN}. Parameters $\lambda_1$ and $\lambda_2$ are determined based on a few pilot runs, each for a few iterations only, to examine the relative contribution of each term in the loss function to the overall loss. Based on these pilot runs, both of these parameters are set to 500. Batch size is set to 10,000, and the learning rate $\alpha$  to 0.003.  A generalized Halton sequence generator algorithm is used to generate N=100,000 collocation points and S=10,000 seeds within the computation domain. Another 100,000 uniformly-distributed boundary collocation points are also generated. The model is trained for 3000 iterations. Figure \ref{E2_contour} shows the accuracy of the PINN solution trained using the proposed approach against the Finite Element solution (using MATLAB Partial Differential Equation Toolbox) in solving a time-dependent diffusion example. The computational time for the Finite Element and PINN results presented in this figure are 32 and 40 seconds, respectively.

\begin{figure}
	\begin{center}
		\begin{subfigure}{0.48 \linewidth}
			\includegraphics[width=0.99\linewidth]{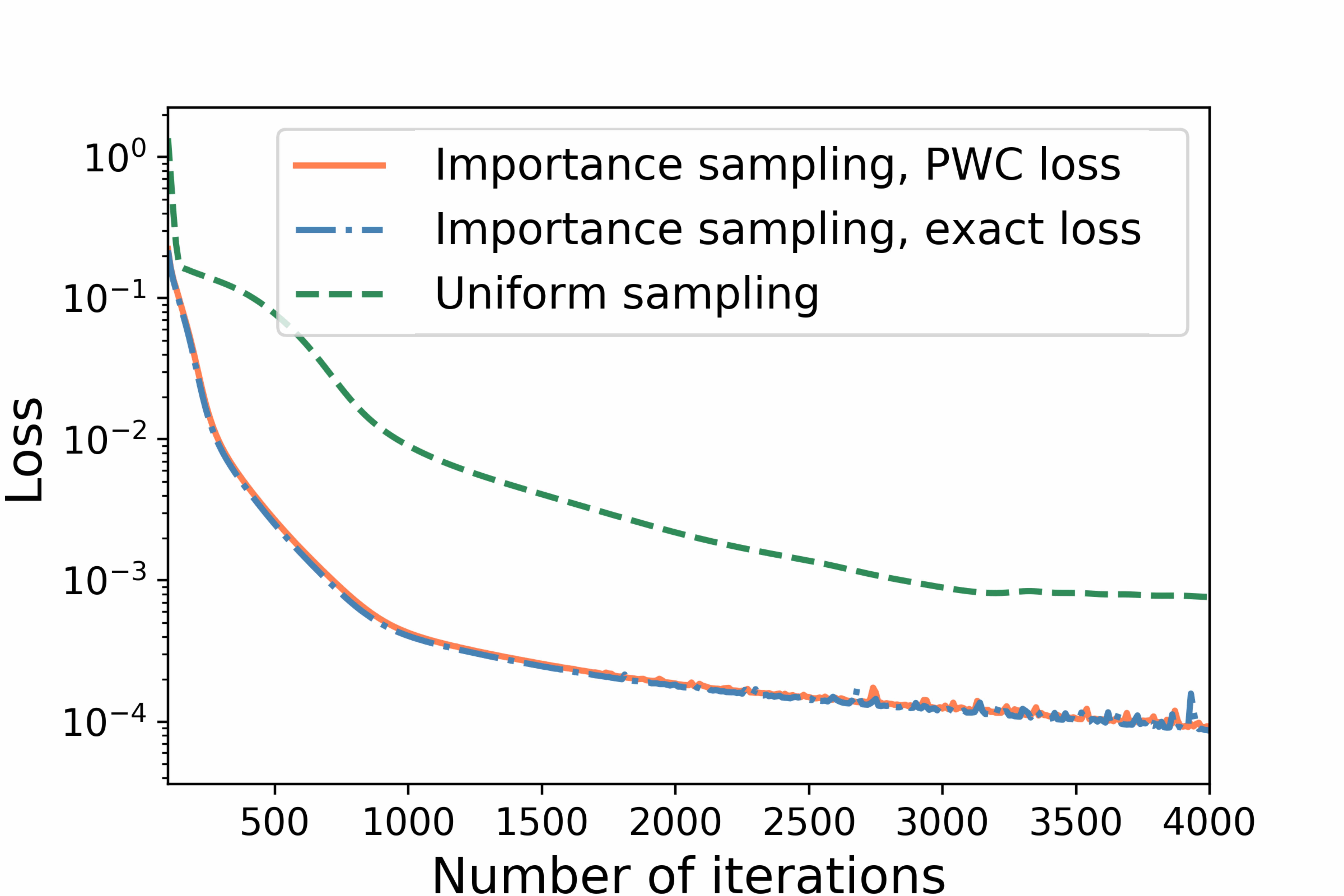}
			\caption{} 
			\label{E2_results.iter}
		\end{subfigure}
		\quad
		\begin{subfigure}{0.48 \linewidth}
			\includegraphics[width=0.99\linewidth]{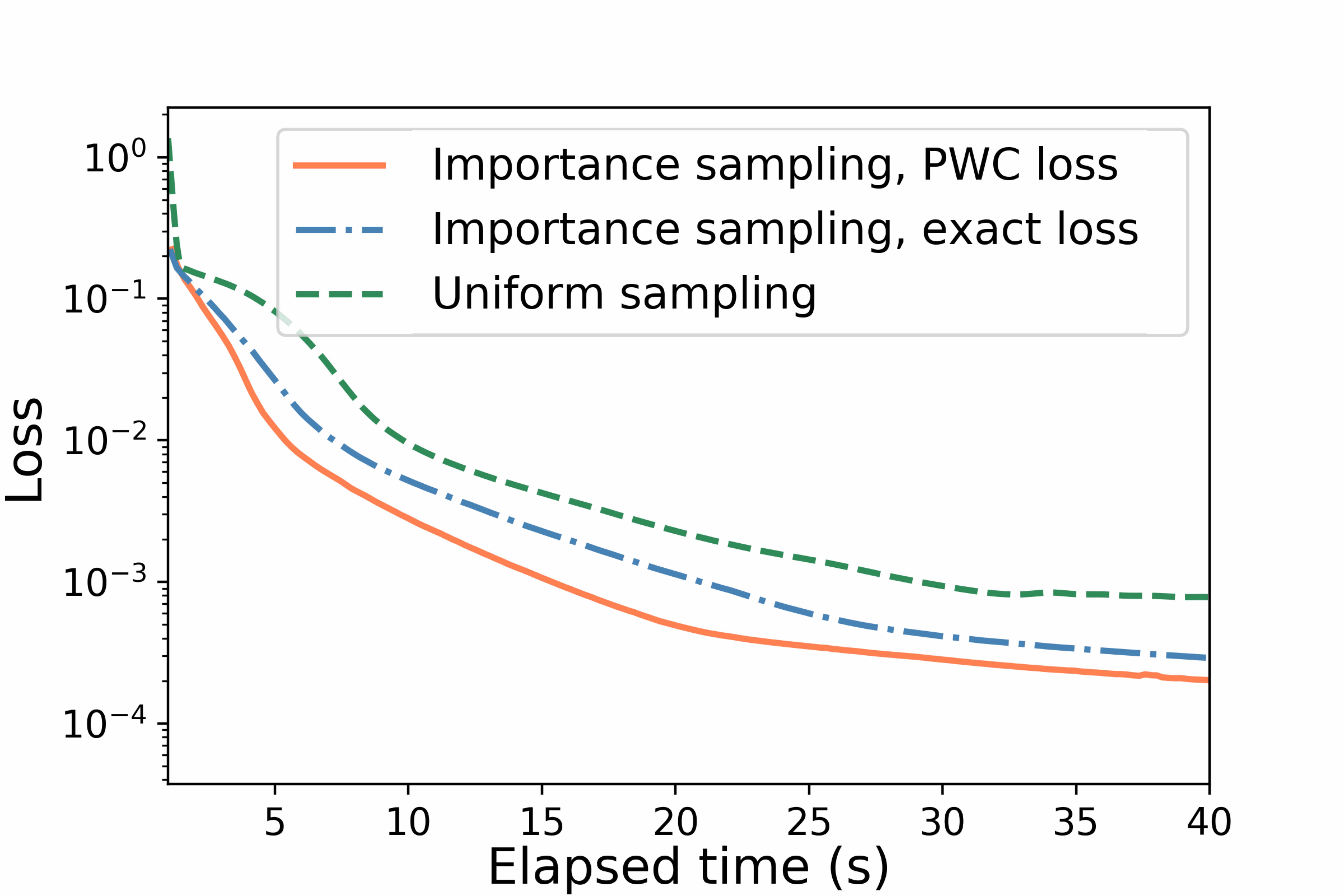}
			\caption{} 
			\label{E2_results.time}
		\end{subfigure}
		\captionsetup{}
	\end{center}
	\caption{A comparison between the performance of the three approaches of uniform sampling, importance sampling with exact loss evaluation, and importance sampling with approximate loss evaluation for training of a PINN solution to transient diffusion defined in Equation \ref{eqn:diffusion}. }
	\label{E2_results}
\end{figure}

Figure \ref{E2_results} compares the computational performance of the importance sampling approach with exact and PWC loss evaluations versus that of the uniform sampling approach, in terms of number of iterations and elapsed time. Once again, it can be seen in Figure~\ref{E2_results.iter} that the PWC approximation to loss is in fact a good approximation. Also, Figure~\label{E2_results.time}   shows that the PWC importance sampling approach provides a better computational efficiency compared to the other two approaches, and numerically demonstrate the effectiveness of this method in accelerating the training of PINNs.

\section{Conclusion}

PINNs are a recently-developed class of machine-learning-based methods that can be used to solve PDEs. Although PINNs offer unique advantages over the existing classical numerical methods for PDE, in their current form they are not expected to dominate the best of the classical methods which  have been substantially improved over the past few decades and are optimized in terms of efficiency and robustness. With that in mind, this paper takes a step forward toward improving the computational efficiency of PINNs for solving PDEs. Specifically, in this paper we presented a new approach for training of PINNs based on importance sampling, which selects training points according to a proposal distribution proportional to a piece-wise constant approximation of the loss function. We showed that this approach does not introduce any new hyperparameters, and is straightforward to implement into the existing PINN codes. With some theoretical evidences and also three numerical examples of elasticity and transient diffusion, we demonstrated the effectiveness of the proposed importance sampling approach in improvising the efficiency of PINN training.

While the theoretical background and numerical results presented in this paper provide sufficient evidence that use of the proposed importance sampling approach is promising for accelerating the convergence of PINN training, it is still necessary to further investigate the success of our proposed importance sampling approach in solving more challenging problems, such as stochastic PDEs and time-dependent PDEs with highly oscillatory or non-monotonic solutions. Also, the following extensions to this work can be addressed in future studies: (1) generalizing the algorithm to distributed importance sampling for even more computational efficiency, where a number of workers search for the most informative collocation points while a single worker updates the model parameters; and (2) investigating the performance of the proposed algorithm in improving the efficiency of PINNs for data-driven PDE discovery.

\bibliography{PINN_IS}

\begin{thebibliography}{10}
\expandafter\ifx\csname url\endcsname\relax
  \def\url#1{\texttt{#1}}\fi
\expandafter\ifx\csname urlprefix\endcsname\relax\def\urlprefix{URL }\fi
\expandafter\ifx\csname href\endcsname\relax
  \def\href#1#2{#2} \def\path#1{#1}\fi

\bibitem{lagaris1998artificial}
I.~E. Lagaris, A.~Likas, D.~I. Fotiadis, Artificial neural networks for solving
  ordinary and partial differential equations, IEEE Transactions on Neural
  Networks 9~(5) (1998) 987--1000.

\bibitem{raissi2019physics}
M.~Raissi, P.~Perdikaris, G.~Karniadakis, Physics-informed neural networks: A
  deep learning framework for solving forward and inverse problems involving
  nonlinear partial differential equations, Journal of Computational Physics
  378 (2019) 686--707.

\bibitem{baydin2018automatic}
A.~G. Baydin, B.~A. Pearlmutter, A.~A. Radul, J.~M. Siskind, Automatic
  differentiation in machine learning: a survey, Journal of Marchine Learning
  Research 18 (2018) 1--43.

\bibitem{bottou2012stochastic}
L.~Bottou, Stochastic gradient descent tricks, in: Neural networks: Tricks of
  the trade, Springer, 2012, pp. 421--436.

\bibitem{dissanayake1994neural}
M.~Dissanayake, N.~Phan-Thien, Neural-network-based approximations for solving
  partial differential equations, communications in Numerical Methods in
  Engineering 10~(3) (1994) 195--201.

\bibitem{psichogios1992hybrid}
D.~C. Psichogios, L.~H. Ungar, A hybrid neural network-first principles
  approach to process modeling, AIChE Journal 38~(10) (1992) 1499--1511.

\bibitem{raissi2017physics}
M.~Raissi, P.~Perdikaris, G.~E. Karniadakis, Physics informed deep learning
  (part i): Data-driven solutions of nonlinear partial differential equations,
  arXiv preprint arXiv:1711.10561.

\bibitem{berg2018unified}
J.~Berg, K.~Nystr{\"o}m, A unified deep artificial neural network approach to
  partial differential equations in complex geometries, Neurocomputing 317
  (2018) 28--41.

\bibitem{sirignano2017dgm}
J.~Sirignano, K.~Spiliopoulos, Dgm: A deep learning algorithm for solving
  partial differential equations, arXiv preprint arXiv:1708.07469.

\bibitem{guo2019deep}
H.~Guo, X.~Zhuang, T.~Rabczuk, A deep collocation method for the bending
  analysis of kirchhoff plate, CMC-COMPUTERS MATERIALS \& CONTINUA 59~(2)
  (2019) 433--456.

\bibitem{weinan2018deep}
E.~Weinan, B.~Yu, The deep ritz method: a deep learning-based numerical
  algorithm for solving variational problems, Communications in Mathematics and
  Statistics 6~(1) (2018) 1--12.

\bibitem{goswami2019transfer}
S.~Goswami, C.~Anitescu, S.~Chakraborty, T.~Rabczuk, Transfer learning enhanced
  physics informed neural network for phase-field modeling of fracture, arXiv
  preprint arXiv:1907.02531.

\bibitem{jagtap2019adaptive}
A.~D. Jagtap, G.~E. Karniadakis, Adaptive activation functions accelerate
  convergence in deep and physics-informed neural networks, arXiv preprint
  arXiv:1906.01170.

\bibitem{raissi2018deep}
M.~Raissi, Deep hidden physics models: Deep learning of nonlinear partial
  differential equations, The Journal of Machine Learning Research 19~(1)
  (2018) 932--955.

\bibitem{qin2018data}
T.~Qin, K.~Wu, D.~Xiu, Data driven governing equations approximation using deep
  neural networks, arXiv preprint arXiv:1811.05537.

\bibitem{long2017pde}
Z.~Long, Y.~Lu, X.~Ma, B.~Dong, Pde-net: Learning pdes from data, arXiv
  preprint arXiv:1710.09668.

\bibitem{nabian2019deep}
M.~A. Nabian, H.~Meidani, A deep learning solution approach for
  high-dimensional random differential equations, Probabilistic Engineering
  Mechanics 57 (2019) 14--25.

\bibitem{raissi2019deep}
M.~Raissi, Z.~Wang, M.~S. Triantafyllou, G.~E. Karniadakis, Deep learning of
  vortex-induced vibrations, Journal of Fluid Mechanics 861 (2019) 119--137.

\bibitem{raissi2018hiddena}
M.~Raissi, A.~Yazdani, G.~E. Karniadakis, Hidden fluid mechanics: A
  navier-stokes informed deep learning framework for assimilating flow
  visualization data, arXiv preprint arXiv:1808.04327.

\bibitem{zhu2019physics}
Y.~Zhu, N.~Zabaras, P.-S. Koutsourelakis, P.~Perdikaris, Physics-constrained
  deep learning for high-dimensional surrogate modeling and uncertainty
  quantification without labeled data, Journal of Computational Physics 394
  (2019) 56--81.

\bibitem{meng2019composite}
X.~Meng, G.~E. Karniadakis, A composite neural network that learns from
  multi-fidelity data: Application to function approximation and inverse pde
  problems, arXiv preprint arXiv:1903.00104.

\bibitem{yang2019adversarial}
Y.~Yang, P.~Perdikaris, Adversarial uncertainty quantification in
  physics-informed neural networks, Journal of Computational Physics 394 (2019)
  136--152.

\bibitem{kissas2019machine}
G.~Kissas, Y.~Yang, E.~Hwuang, W.~R. Witschey, J.~A. Detre, P.~Perdikaris,
  Machine learning in cardiovascular flows modeling: Predicting pulse wave
  propagation from non-invasive clinical measurements using physics-informed
  deep learning, arXiv preprint arXiv:1905.04817.

\bibitem{xu2019neural}
K.~Xu, E.~Darve, The neural network approach to inverse problems in
  differential equations, arXiv preprint arXiv:1901.07758.

\bibitem{yang2018physics}
L.~Yang, D.~Zhang, G.~E. Karniadakis, Physics-informed generative adversarial
  networks for stochastic differential equations, arXiv preprint
  arXiv:1811.02033.

\bibitem{raissi2018forward}
M.~Raissi, Forward-backward stochastic neural networks: Deep learning of
  high-dimensional partial differential equations, arXiv preprint
  arXiv:1804.07010.

\bibitem{beck2018solving}
C.~Beck, S.~Becker, P.~Grohs, N.~Jaafari, A.~Jentzen, Solving stochastic
  differential equations and kolmogorov equations by means of deep learning,
  arXiv preprint arXiv:1806.00421.

\bibitem{weinan2017deep}
E.~Weinan, J.~Han, A.~Jentzen, Deep learning-based numerical methods for
  high-dimensional parabolic partial differential equations and backward
  stochastic differential equations, Communications in Mathematics and
  Statistics (2017) 1--32.

\bibitem{nabian2020physics}
M.~A. Nabian, H.~Meidani, Physics-driven regularization of deep neural networks
  for enhanced engineering design and analysis, Journal of Computing and
  Information Science in Engineering 20~(1).

\bibitem{bottou2010large}
L.~Bottou, Large-scale machine learning with stochastic gradient descent, in:
  Proceedings of COMPSTAT'2010, Springer, 2010, pp. 177--186.

\bibitem{press2007numerical}
W.~H. Press, S.~A. Teukolsky, W.~T. Vetterling, B.~P. Flannery, Numerical
  recipes 3rd edition: The art of scientific computing, Cambridge university
  press, 2007.

\bibitem{katharopoulos2018not}
A.~Katharopoulos, F.~Fleuret, Not all samples are created equal: Deep learning
  with importance sampling, arXiv preprint arXiv:1803.00942.

\bibitem{katharopoulos2017biased}
A.~Katharopoulos, F.~Fleuret, Biased importance sampling for deep neural
  network training, arXiv preprint arXiv:1706.00043.

\bibitem{alain2015variance}
G.~Alain, A.~Lamb, C.~Sankar, A.~Courville, Y.~Bengio, Variance reduction in
  sgd by distributed importance sampling, arXiv preprint arXiv:1511.06481.

\bibitem{marsland2014machine}
S.~Marsland, Machine learning: an algorithmic perspective, Chapman and
  Hall/CRC, 2014.

\bibitem{aurenhammer1991voronoi}
F.~Aurenhammer, Voronoi diagrams—a survey of a fundamental geometric data
  structure, ACM Computing Surveys (CSUR) 23~(3) (1991) 345--405.

\bibitem{lecun2015deep}
Y.~LeCun, Y.~Bengio, G.~Hinton, Deep learning, Nature 521~(7553) (2015)
  436--444.

\bibitem{goodfellow2016deep}
I.~Goodfellow, Y.~Bengio, A.~Courville, Deep learning, MIT press, 2016.

\bibitem{kingma2014adam}
D.~P. Kingma, J.~Ba, Adam: A method for stochastic optimization, arXiv preprint
  arXiv:1412.6980.

\bibitem{duchi2011adaptive}
J.~Duchi, E.~Hazan, Y.~Singer, Adaptive subgradient methods for online learning
  and stochastic optimization, Journal of Machine Learning Research 12~(Jul)
  (2011) 2121--2159.

\bibitem{zeiler2012adadelta}
M.~D. Zeiler, Adadelta: an adaptive learning rate method, arXiv preprint
  arXiv:1212.5701.

\bibitem{sutskever2013importance}
I.~Sutskever, J.~Martens, G.~Dahl, G.~Hinton, On the importance of
  initialization and momentum in deep learning, in: International conference on
  machine learning, 2013, pp. 1139--1147.

\bibitem{bochev2006least}
P.~B. Bochev, M.~D. Gunzburger, Least-squares finite element methods, Springer,
  2006.

\bibitem{ruder2016overview}
S.~Ruder, An overview of gradient descent optimization algorithms, arXiv
  preprint arXiv:1609.04747.

\bibitem{halton1960efficiency}
J.~H. Halton, On the efficiency of certain quasi-random sequences of points in
  evaluating multi-dimensional integrals, Numerische Mathematik 2~(1) (1960)
  84--90.

\bibitem{faure2009generalized}
H.~Faure, C.~Lemieux, Generalized halton sequences in 2008: A comparative
  study, ACM Transactions on Modeling and Computer Simulation (TOMACS) 19~(4)
  (2009) 15.

\bibitem{sobol1967distribution}
I.~M. Sobol', On the distribution of points in a cube and the approximate
  evaluation of integrals, Zhurnal Vychislitel'noi Matematiki i Matematicheskoi
  Fiziki 7~(4) (1967) 784--802.

\bibitem{joe2008constructing}
S.~Joe, F.~Y. Kuo, Constructing sobol sequences with better two-dimensional
  projections, SIAM Journal on Scientific Computing 30~(5) (2008) 2635--2654.

\bibitem{hammersley1960monte}
J.~M. Hammersley, Monte carlo methods for solving multivariable problems,
  Annals of the New York Academy of Sciences 86~(3) (1960) 844--874.

\bibitem{hammersley2013monte}
J.~Hammersley, Monte carlo methods, Springer Science \& Business Media, 2013.

\bibitem{abadi2016tensorflow}
M.~Abadi, P.~Barham, J.~Chen, Z.~Chen, A.~Davis, J.~Dean, M.~Devin,
  S.~Ghemawat, G.~Irving, M.~Isard, et~al., Tensorflow: A system for
  large-scale machine learning, in: 12th $\{$USENIX$\}$ Symposium on Operating
  Systems Design and Implementation ($\{$OSDI$\}$ 16), 2016, pp. 265--283.

\bibitem{chikazawa2001particle}
Y.~Chikazawa, S.~Koshizuka, Y.~Oka, A particle method for elastic and
  visco-plastic structures and fluid-structure interactions, Computational
  Mechanics 27~(2) (2001) 97--106.

\bibitem{TOUSSAINT2017148}
E.~Toussaint, S.~Durif, A.~Bouchaïr, M.~Grédiac, Strain measurements and
  analyses around the bolt holes of structural steel plate connections using
  full-field measurements., Engineering Structures 131 (2017) 148 -- 162.

\bibitem{hennigh2020nvidia}
O.~Hennigh, S.~Narasimhan, M.~A. Nabian, A.~Subramaniam, K.~Tangsali,
  M.~Rietmann, J.~d.~A. Ferrandis, W.~Byeon, Z.~Fang, S.~Choudhry, Nvidia
  simnet\^{}$\{$TM$\}$: an ai-accelerated multi-physics simulation framework,
  arXiv preprint arXiv:2012.07938.

\bibitem{simnet2020}
O.~Hennigh, K.~Tangsali, A.~Subramaniam, S.~Narasimhan, M.~Nabian, J.~d.~A.
  Ferrandis, S.~Choudhry, Simnet: A neural network solver for multi-physics
  applications., Bulletin of the American Physical Society.

\end{thebibliography}

\end{document}